\documentclass[11pt]{article}
\usepackage{times}
\usepackage{color}

\usepackage{bm}
\usepackage{amssymb}

\usepackage[authoryear]{natbib}
\usepackage{rotating}
\usepackage{bbm}
\usepackage{latexsym}
\usepackage{fancyhdr}
\usepackage[ruled]{algorithm2e}
\usepackage{graphicx}
\usepackage{url}
\usepackage{subfig}
\usepackage{amsmath}
\usepackage{amsthm}
\usepackage{multirow}
\usepackage[capitalize]{cleveref}
\newtheorem{definition}{Definition}
\crefname{definition}{Definition}{Definitions}
\newtheorem{theorem}{Theorem}
\crefname{theorem}{Theorem}{Theorems}

\crefname{proposition}{Proposition}{Propositions}
\newtheorem{corollary}{Corollary}

\newtheorem{lemma}{Lemma}
\newtheorem{assumption}{Assumption}
\crefname{assumption}{Assumption}{Assumptions}

\allowdisplaybreaks  % allow new page in a math environment
\usepackage{amsfonts}
\usepackage{amssymb}
\newcommand{\Real}[0]{\ensuremath{\mathbb{R}}}

\newcommand{\Beta}[0]{\ensuremath{\mathcal{B}}}
\newcommand{\lconst}[0]{\ensuremath{c_\text{l}}}
\newcommand{\uconst}[0]{\ensuremath{c_\text{u}}}
\newcommand{\priX}[0]{\ensuremath{\mathbf{Z}^{(\mathbf{X})}}}
\newcommand{\prix}[0]{\ensuremath{\mathbf{z}^{(\mathbf{X})}}}

\newcommand{\prixj}[0]{\ensuremath{z_j^{(\mathbf{X})}}}
\newcommand{\priXi}[0]{\ensuremath{\mathbf{Z}_i^{(\mathbf{X})}}}
\newcommand{\prixi}[0]{\ensuremath{\mathbf{z}_i^{(\mathbf{X})}}}

\newcommand{\prixij}[0]{\ensuremath{z_{ij}^{(\mathbf{X})}}}
\newcommand{\priY}[0]{\ensuremath{Z^{(Y)}}}
\newcommand{\priy}[0]{\ensuremath{z^{(Y)}}}
\newcommand{\priYi}[0]{\ensuremath{Z_i^{(Y)}}}
\newcommand{\priyi}[0]{\ensuremath{z_i^{(Y)}}}

\newcommand{\Expe}[2]{\ensuremath{\mathbb{E}_{#1}\left[#2\right]}}
\newcommand{\ip}[2]{\ensuremath{{#1}^{\top} {#2}}}

\newcommand{\indicator}[1]{\mathbf{1}\left[#1\right]}

\newcommand{\checkl}{\ensuremath{\rho}}
\newcommand{\bitf}[0]{\ensuremath{{Q}_{\text{bf}}}}

\newcommand{\argmax}{\mathop{\rm arg~max}\limits}

\newcommand{\captionfonts}{\normalsize}

\makeatletter  
\long\def\@makecaption#1#2{%
  \vskip\abovecaptionskip
  \sbox\@tempboxa{{\captionfonts #1: #2}}%
  \ifdim \wd\@tempboxa >\hsize
    {\captionfonts #1: #2\par}
  \else
    \hbox to\hsize{\hfil\box\@tempboxa\hfil}%
  \fi
  \vskip\belowcaptionskip}
\makeatother   
\begin{document}
\hspace{13.9cm}1

\ \vspace{20mm}\\

{\begin{center}
{\LARGE One-bit Submission for Locally Private Quasi-MLE: Its Asymptotic Normality and Limitation}
\end{center}

\ \\
{\bf \large 
Hajime Ono$^{1}$,
Kazuhiro Minami$^{1,2}$,
Hideitsu Hino$^{1,2,3}$
}

\begin{flushleft}
$^1$ The Graduate University for Advanced Studies (SOKENDAI),\\
$^2$ The Institute of Statistical Mathematics\\
$^3$ RIKEN AIP
\end{flushleft}}
%\ \\[-2mm]
%{\bf Keywords: 
%graph structure estimation, graph Laplacian, random walk, diffusion kernel}
\thispagestyle{fancy}
\rhead{}
\lhead{}
\chead{Draft of the paper to appear in \lq\lq AISTATS2022 \rq\rq}
%\thispagestyle{myheadings}
%\markboth{}{aho}
%%
%Abstract
\begin{center} {\bf Abstract} \end{center}
Local differential privacy~(LDP) is an information-theoretic privacy definition suitable for statistical surveys that involve an untrusted data curator.
An LDP version of quasi-maximum likelihood estimator~(QMLE) has been developed, but the existing method to build LDP QMLE is difficult to implement for a large-scale survey system in the real world due to long waiting time, expensive communication cost, and the boundedness assumption of derivative of a log-likelihood function.
We provided an alternative LDP protocol without those issues, which is potentially much easily deployable to a large-scale survey. 
We also provided sufficient conditions for the consistency and asymptotic normality and limitations of our protocol.
Our protocol is less burdensome for the users, and the theoretical guarantees cover more realistic cases than those for the existing method.
%%%%%%%%%%%

\section{INTRODUCTION}
The collection and use of data related to individuals continue at an unprecedented pace, raising a critical question: How do we balance the benefits of data use with the inherent privacy risks involved?
One option is privacy protection based on differential privacy~(DP)\citep{Dwork2006cali,Dwork2014foundation}, whose information-theoretic definition requires data curators such as IT companies to stochastically perturb the results of research before making them available to third parties or the public.
DP statistical data processing has been widely studied both theoretically~\citep{Bassily2014erm} and empirically~\citep{Abadi2016deepdp}.
However, protection with DP does not work when the curator is adversarial.
In fact, IT companies sometimes betray their users~\citep{Day2019amazon}.

To ensure that user privacy is protected even if the company is adversarial, local differential privacy~(LDP)\citep{Kasiviswanathan2011nldp,Duchi2013minimax} can be employed.
By definition, LDP requires that the users themselves stochastically perturb their sensitive records before providing the records to a company.
This perturbation ensures that no one can deterministically know the records.
Notably, Google and Apple have conducted statistical surveys that guarantee user privacy based on this definition~\citep{Erlingsson2014rappor,Apple2014scale}.

LDP versions of many statistical tools have been developed, including heavy-hitter estimation~\citep{Erlingsson2014rappor,Fanti2016rappor,Bassily2015succinct,Qin2016heavy}, discrete distribution estimation~\citep{Kairouz2016discrete,Ding2017telemetry}, t-tests~\citep{Ding2018t-test}, chi-squared tests ~\citep{Gaboardi2018chi-square} and sparse linear regression~\citep{Wang2019sparse}.
An LDP quasi-maximum likelihood estimator (QMLE) can also be included among these tools.
QMLE is an estimator of a parameter likely approximating a distribution $F$ generating a set of observations $D_n=\{\mathbf{x}_1,\hdots,\mathbf{x}_n\}$, from model family $\{F_\theta:\theta\in\Theta\}$.
The likelihood of parameter $\theta$ is evaluated using the log-likelihood function $\ell(\theta;D_n)=\sum_{i=1}^n\log f_\theta(\mathbf{x}_i)/n$, and 
QMLE $\hat{\theta}_n$ is defined as the maximizer of $\ell(\theta;D_n)$.  
MLE is a special case in which there is a correct model:  $F\in\{F_\theta:\theta\in\Theta\}$.
Since no one observes the raw data under the LDP constraint, it is too optimistic to assume that we can specify a family including the true distribution.
In this paper, we mainly consider QMLE rather than MLE.
Under regularity conditions, QMLE has asymptotic normality.
By understanding its normality, the curator is able to determine how likely and by how much the estimator is to deviate from the optimal point.
Moreover, with the asymptotic normality, we can perform the Wald test, which is an important application~\citep{Vaart2000asymptotic}.

\citet{Bhowmick2018protection} provided a framework for LDP M-estimators, which is a superclass of LDP QMLEs. It approximates the maximizer of an objective function with stochastic gradient descent. They showed that the covariance matrix of the normal distribution on which the estimator converges agrees with minimax optimal ones up to a constant.

However, the existing  protocol may be difficult to deploy for a large-scale system in the real world due to the following three problems:
(i) it requires a long waiting time for users, (ii) it is communication inefficient, and (iii) it requires finiteness of the derivative of the objective function.
The existing protocol is interactive wherein the communication of the $i$th user depends on those of the previous $i-1$ users. Though this interactivity gives more accurate statistics~\citep{Smith2017interaction}, it causes a long waiting time for users when millions of users are involved in the protocol.
Communication efficiency is a non-ignorable problem for large-scale implementation, especially on Edge or IoT devices.
When the parameter is $d$-dimensional and each component of the parameter uses float as a data type, each user submits $32d$ bits.

It is also of great practical importance to be able to apply to unbounded domain data. 
The LDP constraints require a user to perturb her record so as to be indistinguishable from the other candidate records in the domain.
An unbounded domain makes it difficult to satisfy this requirement since no one knows how many candidate values exist in the domain.

We provide low-user-side-cost protocols that involve no waiting time, require no boundedness assumption, and avoid high communication costs for QMLEs of regression.
In this paper, we focus on regression which is a wide and important class.
To eliminate waiting time, we abandon interactivity.
Although less accurate than interactive methods, our protocol has a significant advantage in that the execution time on the user side is constant regardless of the number of users.
To remove the boundedness assumption, we incorporate truncation into the protocol.
This simple technique makes it possible for the protocol to perform safely even when the record domain is unbounded.
For communication efficiency, we adopt the one-bit submission strategy whereby a record is stochastically quantized into a binary value~\citep{McGregor2010two-party,Seide2014one_bit,Bassily2015succinct,Ding2018t-test,Wang2018erm}.
This strategy significantly reduces the communication cost.
See \cref{tab:communication} for a quick comparison of the communication costs and waiting time.

\begin{table}[b]
    \centering
    \caption{Comparison of communication costs in number of submitting bits and waiting time of the protocols of the existing protocol~\citep{Bhowmick2018protection} and our protocol in two scenarios where explanatory variables $\mathbf{X}$ are public and private. $d$ is the dimension of parameter, $k$ is number of explanatory variables, and $n$ is the number of users.}
    \label{tab:communication}
    \begin{tabular}{c|c|c|c|c}
        Id & Scenario & Server & User & Wait\\
        \hline\hline
        \multirow{2}{*}{Bhowmick2018} & $\mathbf{X}$ pub & $32k$ & $32d$ & \multirow{2}{*}{$O(n)$}\\
        & $\mathbf{X}$ pri & $0$ & $32k$\\\hline
        \multirow{2}{*}{Ours} & $\mathbf{X}$ pub & $32k$ & $1$ & \multirow{2}{*}{$O(1)$}\\
        & $\mathbf{X}$ pri & $0$ & $d+1$ 
    \end{tabular}
\end{table}

As the main contributions of this paper, (i) we give consistency and asymptotic normality theorem with their sufficient conditions for our QMLEs, and (ii) we make explicit the limitations of the scope of our theoretical analysis.
The asymptotic normality is useful for curators to adequately decide sample size $n$ and privacy parameter $\epsilon$.
The sufficient conditions for our consistent and normality theorems are conditions on the model family and the true distribution.
The curator should check the conditions for the model family when selecting the family.
On the other hand, no one can evaluate the conditions on the true distribution.
We recommend that the curator should carefully consider these conditions with the help of experts.

To discuss the sufficient conditions for our theorems on a concrete problem, we consider $\alpha$-quantile linear regression~\citep{Davino2013quantile}.
With this example, we can see that it is not so difficult to make a model family satisfying the conditions.
Given $\alpha\in(0, 1)$, coefficients estimation for $\alpha$-quantile regression is one of the standard statistical data analyses and QMLE is one of the solutions.
For explanatory variables $\mathbf{X}$ on $\Real^d$ and objective variable $Y$ on $\Real$, the goal of the $\alpha$-quantile regression is to find coefficient $\beta\in\mathcal{B}\subset\Real^d$ such that the inner product $\ip{\beta}{\mathbf{X}}$ well approximates the $\alpha$-quantile of the distribution of $Y$, i.e., $\inf\{y|\Pr(Y\leq y|\mathbf{X})>\alpha\}$. If we consider asymmetric Laplace distributions as the model family, this problem is a likelihood-maximizing  problem.
With this example, we are able to confirm that the conditions regarding the model family are easily satisfied.
In addition, using real data, we observe the asymptotic behavior of our QMLE.
The observations imply that the Frobenius norm of empirical covariance matrix shrinks in proportion to $1/n$ as expected in the asymptotic normality theorem.

We mention some related works.
LDP regression by non-interactive algorithms has been studied in the context of LDP empirical risk minimization e.g., \citep{Smith2017interaction,Zheng2017non-interactive,Wang2018erm,Wang2019non_interactive_linear,Wang2021glm}.
Their targets are not analyses of asymptotic normality but seeking smaller risk.
The studies for non-local differentially privately M-estimators took different ways from us~\citep{Smith2011converge,Chaudhuri2012converge,Avella-Medina2020robust}.
Due to the difference in the privacy models, we do not compare our results with theirs.
\citet{Bhowmick2018protection} showed asymptotic normality of their estimator relying on \citet{Polyak1992acceleration} 's asymptotic-normality proof for the estimators obtained by stochastic gradient descent.
Since we do not use stochastic gradient descent, we prove our theorem by a different method.

The remainder of the paper is organized as follows:
In~\Cref{sec:preliminaries}, we introduce the notation used in this paper and some of the basic concepts. 
In~\Cref{sec:framework}, we describe our protocols for building QMLEs.
In~\Cref{sec:example}, we discuss QMLE for $\alpha$-quantile regression as an illustrative application of the protocol.
In~\Cref{sec:numerical}, we report the results of a numerical experiment with real data. In \Cref{sec:conclusion}, we offer concluding remarks.

\section{PRELIMINARIES}\label{sec:preliminaries}
   
We begin by defining some of the notation used in the paper. We denote by $0_d$ the $d$-length zero vector.
When we take expectation while emphasizing the distribution $F$, we use $F g = \Expe{X\sim F}{g(X)}$ where $g$ is a function. A comprehensive summary of our notation is provided in \cref{app:notation}.

\subsection{Local Differential Privacy}

Local differential privacy is a rigorous privacy definition for distributed statistical analyses.
The definition requires each user to protect her sensitive record individually by stochastic perturbation.
In particular, we consider the case in which users receive no feedback from the curator.
LDP in such a situation is called non-interactive LDP; in this paper, we refer to non-interactive LDP simply as LDP.

We can now formally define LDP.
Assume there are $n$ users, each of whom possesses a sensitive record $R_i$ for $i=1,\cdots,n$.
Let $\mathcal{R}$ be the domain of the records. Assume that there is also a curator who will perform a statistical analysis on the users' records and that each user will submit her perturbed record to the curator.
We can define the perturbation as a conditional distribution $Q(\cdot|R=r)$ and LDP as a property of $Q$.
\begin{definition}[$\epsilon$-LDP]
    Given $\epsilon>0$, distribution $Q$ is $\epsilon$-locally differentially private if, for any $r,r'\in \mathcal{R}$, 
    \begin{align*}
        \sup_{\mathcal{S}\in\sigma(\mathcal{X})}Q(S|R=r)\leq e^\epsilon Q(S|R=r'), 
    \end{align*}
    where $\sigma(\mathcal{X})$ is a $\sigma$-algebra on $\mathcal{X}$.
\end{definition}
This definition requires that the conditional distributions $Q(\cdot|r)$ and $Q(\cdot|r')$ are not so different from each other for any pair $r, r'$ of records in $\mathcal{R}$.
The $\epsilon$ represents the similarity of the conditional distributions.
A smaller $\epsilon$ implies stricter privacy protection but less information of the outputs.
$\epsilon$ thus controls the trade-off between privacy protection and utility.

This paper uses the bit flip~\citep{Ding2018t-test} for the concrete implementation of conditional distribution $Q$ satisfying $\epsilon$-LDP.
The bit flip stochastically maps a finite continuous interval $[\lconst, \uconst]$, where $\lconst$ and $\uconst$ are some real constants such that $\lconst< \uconst$, into discrete binary values $\{z_-, z_+\}$.
Then, for any input $v\in[\lconst, \uconst]$ and with $C_\epsilon=\frac{e^\epsilon+1}{e^\epsilon-1}$, the bit flip is defined as
\begin{align*}
    \bitf(Z=z|v) = 
     \begin{cases}
        \frac{1}{2} - \frac{v-\frac{\uconst+\lconst}{2}}{(\uconst-\lconst)C_\epsilon} & \text{if}~ z=z_-,\\
        \frac{1}{2}+\frac{v-\frac{\uconst+\lconst}{2}}{(\uconst-\lconst)C_\epsilon} & \text{if}~ z=z_+.
     \end{cases}
\end{align*}
When the input is close to $\uconst$, the output is likely to be $z_+$; conversely, when the input is close to $\lconst$, the output is likely to be $z_-$.

\subsection{Quasi-Maximum Likelihood Estimator}

Given observations $D_n=\{\mathbf{x}_1,\hdots,\mathbf{x}_n\}$ generated by distribution $F$, the likelihood of parameter $\theta$ of a model $F_{\theta}$ is evaluated by the log-likelihood function 
\[
    \ell(\theta;D_n)=\frac{1}{n}\sum_{i=1}^n\log f_\theta(\mathbf{x}_i),
\]
where $f_\theta$ is the density function of $F_\theta$.
Roughly speaking, the log-likelihood is the log of the probability that the observations are obtained assuming they are sampled from $F_\theta$.
For the likelihood function, QMLE $\hat\theta_n$ is defined as $\hat{\theta}_n=\text{argmax}_{\theta\in\Theta}\ell(\theta;D_n)$.
Not only $D_n$ but also $\hat{\theta}_n$ itself is a random variable.

In this subsection, we review the consistency and asymptotic normality theorems of QMLEs by \citet{White1982misspecified}. 
To define the log-likelihood function well, we first need to make some assumptions.
The first is that the observations are independently generated from a distribution $F$ and that $F$ has a regular Radon--Nikodym density function $f$.
The second condition requires that the model family also has regular density functions.
\begin{assumption}
    \label{asm:A1}
    Let $\mathcal{\nu}$ be an appropriate measure on $\mathcal{X}$.
    For a constant $k$, the independent $1\times k$ random vectors $\mathbf{X}_i, i = 1,\cdots, n$, have common joint distribution function $F$ on $\mathcal{X}$, a measurable Euclidean space, with measurable Radon--Nikodym density $f = dF/d\nu$.
\end{assumption}
\begin{assumption}
    \label{asm:A2}
    The family of distribution functions $F_\theta(\mathbf{x})$ has Radon--Nikodym densities $f_\theta(\mathbf{x}) = dF_\theta(\mathbf{x})/d\nu$ which are measurable in $x$ for every $\theta\in\Theta$, a compact subset of a $d$-dimensional Euclidean space, and continuous in $\theta$ for every $\mathbf{x}\in\mathcal{X}$.
\end{assumption}

To guarantee consistency, we introduce an additional technical assumption. 
\begin{assumption}
    \label{asm:A3}
    (a) $F\log f$ exists, and $|\log f_\theta(\mathbf{x})| \leq h(\mathbf{x})$ for all $\theta\in\Theta$, where $h$ is integrable with respect to $F$; (b) $F\log f_\theta$ has a unique maximum at $\theta^* \in\Theta$.
\end{assumption}
Under these regularity conditions, the QMLE converges to $\theta^*=\text{argmax}_{\theta\in\Theta}F\log f_\theta$.
\begin{theorem}[Theorem 2.2 in \citep{White1982misspecified}]
\label{thm:miss_consistent}
    Given \cref{asm:A1,asm:A2,asm:A3}, $\hat{\theta}_n\rightarrow \theta^*$ as $n\rightarrow\infty$ for almost every sequence $\{\mathbf{X}_i\}_{i=1}^n$.
\end{theorem}

We also have asymptotic normality under some additional assumptions regarding the existence of scores $\partial \log f_{\theta}(\mathbf{x})/\partial \theta$ and related quantities.
\begin{assumption}\label{asm:A4}
    $\partial \log f_\theta(\mathbf{x})/\partial\theta_j, j = 1, . .. , d$, are measurable of $\mathbf{x}$ for each $\theta\in\Theta$ and continuously differentiable functions of $\theta$ for each $\mathbf{x}\in\mathcal{X}$.
\end{assumption}
\begin{assumption}\label{asm:A5}
    $|\partial \log f_\theta(\mathbf{x})/\partial \theta_{j_1}\cdot\partial \log f_\theta(\mathbf{x})/\partial\theta_{j_2}|$ and $|\partial^2 \log f_\theta(\mathbf{x})/\partial\theta_{j_1}\partial\theta_{j_2}|$, for $j_1, j_2 = 1,\dots, d$ are dominated by functions integrable with respect to $F$ for all $\mathbf{x}$ in $\mathcal{X}$ and $\theta$ in $\Theta$.
\end{assumption}
\begin{assumption}\label{asm:A6}
    (a) $\theta^*$ is interior to $\Theta$;
    (b) $B(\theta)=(F(\partial \log f_\theta/\partial\theta)(\partial \log f_\theta/\partial\theta)^\top$ is nonsingular at $\theta=\theta^*$;
    (c) $\theta^*$ is a regular point of $A(\theta)= F\partial^2\log f_\theta/\partial\theta^2$.
\end{assumption}

The following shows the asymptotic normality.
\begin{theorem}[Theorem 3.2 in \citep{White1982misspecified}]
    \label{thm:miss_normal}
    Given \cref{asm:A1,asm:A2,asm:A3,asm:A4,asm:A5,asm:A6}, 
    \begin{align*}
        & \sqrt{n}(\hat{\theta}_n-\theta^*)\to\mathcal{N}(0, C(\theta^*))
        \quad\\
        \text{where}
        \quad
        & C(\theta) = A(\theta)^{-1}B(\theta)A(\theta)^{-1}.
    \end{align*}
    % Moreover, $C_n(\hat{\theta}_n)\stackrel{\text{a.s.}}{\rightarrow} C(\theta_*)$, element by element.
\end{theorem}
When $F_{\theta^*}=F$, $C(\theta^*)$ is called the Fisher information matrix.

\subsection{Quantile Regression}
\label{ssec:quantile}

Linear quantile regression deals with the statistical problem of finding coefficients $\beta\in\mathcal{B} \subset \mathbb{R}^d$ such that, given $\mathbf{x}$, the inner product $\ip{\beta}{\mathbf{x}}$ well approximates the $\alpha$-quantile $\inf\{y|F(Y\leq y|\mathbf{x})>\alpha\}$ of $Y|\mathbf{x}$.
The problem is often formulated as an optimization problem finding $\beta\in\mathcal{B}$ that minimizes the following objective function: Given observations $\{\mathbf{x}_i, y_i\}_{i=1}^n$, 
\begin{align}
  \sum_{i=1}^n \checkl_\alpha(y_i-\ip{\beta}{\mathbf{x}_i})
  \text{ where }
  \checkl_\alpha(\tau) = 
  \begin{cases}
    (\alpha - 1) \tau & \text{if}\hspace{2pt} \tau \leq 0,\\
    \alpha \tau & \text{if}\hspace{2pt} \tau > 0.
  \end{cases}
  \label{eq:problem}
\end{align}
$\checkl_\alpha$ is a convex function, which is called the check loss.

If we assume that objective variable $Y$ is sampled from the asymmetric Laplace distribution defined below, the minimization of (\ref{eq:problem}) is equivalent to the likelihood maximization for the parameter of the distributions: With $\sigma > 0$, 
\begin{align}
    f_Y(y;\alpha, \mu, \sigma)=\frac{\alpha(1-\alpha)}{\sigma} \exp\left(-\checkl_\alpha\left(\frac{y-\mu}{\sigma}\right)\right).
    \label{eq:asymmetric_laplace_distribution}
\end{align}
Hence the log-likelihood function is written as
\begin{align}
    \nonumber
    & \frac{1}{n}\sum_{i=1}^n \log f_{Y}(y_i; \alpha,\ip{\beta}{\mathbf{x}_i},\sigma)\\
    = & \log\frac{\alpha(1-\alpha)}{\sigma} - \frac{1}{n\sigma} \sum_{i=1}^n\checkl_\alpha\left(y_i-\ip{\beta}{\mathbf{x}_i}\right)
    \label{eq:non_private_log_likelihood}.
\end{align}

Finally, we revisit the classical result of the asymptotic normality of the MLE.
Let $\hat{\beta}_n\in\Beta$ be the MLE that minimizes (\ref{eq:non_private_log_likelihood}), and let $\beta^*$ be the coefficient such that $F(Y\leq y|\mathbf{X}=\mathbf{x})=F_Y(y;\alpha,\ip{\beta^*}{\mathbf{x}},\sigma)$ for almost every $\mathbf{x}$ and $y$ with appropriate $\alpha$ and $\sigma$.
Then, the sequence of MLEs $\{\hat{\beta}_n\}_n$ converges as
\begin{align}
    \sqrt{n}(\hat{\beta}_n-\beta^*)
    \rightarrow 
    \mathcal{N}(0_d, I^{-1}),
    \label{eq:non_private_normal}
\end{align}
where $\mathcal{N}(0_d, I^{-1})$ is the normal distribution whose mean and covariance are $0_d$ and $I^{-1}$, respectively~\citep{Davino2013quantile}. 
Assuming that $\Expe{}{\mathbf{X}\mathbf{X}^\top}$ is non-singular, $I$ is the Fisher information matrix defined as
\begin{align}
    I 
    = \frac{\alpha(1-\alpha)}{\sigma^2}\Expe{}{\mathbf{X}\mathbf{X}^\top}.\label{eq:Fisher_non_private}
\end{align}

\section{PROPOSED PROTOCOL}
\label{sec:framework}

We provide two protocols for building QMLEs of regression in two different privacy scenarios and give their asymptotic normality theorem.
Then, we remark on their advantages, limitations, and possible future works.

\subsection{Regression with Public $\mathbf{X}$}
\label{ssec:public_X}

In this subsection, we consider regression with sensitive objective variable $Y$ and public explanatory variables $\mathbf{X}$.
This situation may seem strange, but we will give a practical use case.
Consider a situation in which a company is planning to conduct a customer opinion survey on a new product.
The company can control its features set $\mathbf{X}$ and gives a new product with certain features $\mathbf{X}=\mathbf{x}$ to each customer.
The customer gives an evaluation $Y$ for $\mathbf{X}=\mathbf{x}$.
The target of the company is to understand the conditional distribution of $Y$.
In the survey, the company knows the $\mathbf{X}$s and their distribution, and they are public.

The system model is as follows:
There are a single curator and $n$ users.
The curator selects distribution $F_\mathbf{X}$ on $\mathcal{X}\subset\Real^k$, a measurable Euclidean space, generates $\mathbf{X}_i$ for each user $i=1,\cdots,n$ following $F_\mathbf{X}$, and passes them to each user.
Given $\mathbf{X}_i=\mathbf{x}_i$, user $i$ independently generates $Y_i$ following unknown conditional distribution $F(\cdot|\mathbf{x}_i)$ on $\mathcal{Y}\subset\Real$, a measurable space, and truncates it into interval $[\lconst, \uconst]$.

Let $\bar{Y}_i$ be the truncated version of $Y_i$:
\begin{align}
    \bar{Y}_i = t(Y_i) \equiv
    \begin{cases}
        \lconst & \text{if } Y_i \leq \lconst,\\
        Y_i & \text{if } \lconst < Y_i < \uconst,\\
        \uconst & \text{if } Y_i\geq \uconst.
    \end{cases}
    \label{eq:truncated}
\end{align}
We let $\bar{y}_i$ be a realization of $\bar{Y}_i$.
Then, the user perturbs $\bar{y}_i$ by the bit flip.
$Z_i$ that is perturbed $\bar{Y}_i$ distributes as 
\begin{align}
    p(Z_i=z|\mathbf{X}_i=\mathbf{x})
    = \int \bitf(z|t(y)) dF(y|\mathbf{x}).\label{eq:p_public}
\end{align}
User $i$ submits $z_i$ which is a realization of $Z_i$ to the curator.
The user submission is always only one bit.

The curator considers model family $\{F_\beta(\cdot|\mathbf{x}):\beta\in\Beta, \mathbf{x}\in\mathcal{X}\}$ that consists of conditional distributions parameterized by $\Beta$, a compact subset of a $d$-dimensional Euclidean space.
For each $\beta\in\Beta$, we define conditional density function $p_\beta(z|\mathbf{x})$ by replacing $F$ by $F_\beta$ in (\ref{eq:p_public}).
The target of the curator is to find $\beta$ such that $P_\beta$ well approximates $P$.

In this subsection, we write $P$ and $P_\beta$ to designate joint distributions $P(\mathbf{x}, z)$ and $P_\beta(\mathbf{x}, z)$ rather than conditional distributions $P(z|\mathbf{x})$ and $P_\beta(z|\mathbf{x})$.

Given observations $D_n=\{(z_i, \mathbf{x}_i)\}_{i=1}^n$, the log-likelihood function is defined as 
\begin{align*}
    & \ell(\beta;D_n) \equiv \frac{1}{n}\sum_{i=1}^n \log p_\beta(\mathbf{x}_i, z_i)\\
    = & \frac{1}{n}\sum_{i=1}^n (z_i \log \Lambda_\epsilon(\beta,\mathbf{x}_i)
    + (1-z_i)\log(1-\Lambda_\epsilon(\beta, \mathbf{x}_i))\\
    & \hspace{20pt} + \log F_\mathbf{X}(\mathbf{x}_i))
\end{align*}
where $\Lambda_\epsilon(\beta, \mathbf{x}) = p_\beta(z=1|\mathbf{x})$.
We define $\hat{\beta}_n=\text{arg max}_{\beta\in\Beta}\ell(\beta;D_n)$ and $\beta^*=\text{arg max}_{\beta\in\Beta}P\log p_\beta$.
The model selection and optimization are performed by the curator, and the users do not have to care about them.
The curator can change hyperparameters excepting $\uconst,\lconst$ and $\epsilon$ and can try multiple model families without any additional cost for the users.
The pseudo-code is included in \cref{app:pseudo}.

Now, we analyze the behavior of $\hat{\beta}_n$. 
To derive the consistency of our QMLE, we replace $F$ and $F_\theta$ in \cref{thm:miss_consistent} with $P$ and $P_\beta$, respectively.
We find the conditions under which \cref{asm:A1,asm:A2,asm:A3} are satisfied while replacing $F$ and $F_\theta$ with $P$ and $P_\beta$. To satisfy \cref{asm:A1,asm:A2}, we introduce the following assumptions.
\begin{assumption}\label{asm:fyx_measurable}
    Conditional distribution $F(\cdot|\mathbf{x})$ has a Radon--Nikodym density function $f(y|\mathbf{x})=dF(y|\mathbf{x})/d\nu$ which is measurable in $y$ for every $\mathbf{x}\in\mathcal{X}$.
\end{assumption}
\begin{assumption}
    \label{asm:fx_measurable}
    $F_\mathbf{X}$ has a measurable Radon-Nikodym density $f_\mathbf{X}=dF_\mathbf{X}/d\mu$ with some appropriate measure $\mu$.
\end{assumption}
\begin{assumption}
    \label{asm:fbetayx_measurable}
    The family of distribution functions $F_\beta(y|\mathbf{x})$ has Radon--Nikodym densities $f_\beta(y|\mathbf{x}) = dF_\beta(y|\mathbf{x})/d\nu$ which are measurable in $y$ for every $\mathbf{x}\in\mathcal{X}$ and $\beta\in\Beta$, and continuous in $\beta$ for every $\mathbf{x}\in\mathcal{X}$ and $y\in\mathcal{Y}$.
\end{assumption}
These assumptions are satisfied with many distributions e.g., Gaussian and Bernoulli distributions.
From these assumptions, it is obvious that $P(\mathbf{x}, z), P_\beta(\mathbf{x}, z)$ are measurable and that the density functions $p(\mathbf{x}, z)=p(z|\mathbf{x})f(\mathbf{x})$ and $p_\beta(\mathbf{x}, z)=p_\beta(z| \mathbf{x})f(\mathbf{x})$ exist.

In order for the QMLE for regression parameter to satisfy \cref{asm:A3}, we consider the following two conditions.
The first one is the existence of $P\log p$ and integrable function $h(\mathbf{x}, z)$ such that $|\log p_\beta(\mathbf{x}, z)|\leq h(\mathbf{x}, z)$ for all $\beta\in\Beta$.
$P\log p$ can be extended as
\[
    P\log p = F_\mathbf{X} (P_{\cdot|\mathbf{X}} \log p(\cdot|\mathbf{X}) + \log f_\mathbf{X}(\mathbf{X})).
\]
Since $\log p(\cdot|\mathbf{X})$ is always bounded away from $-\infty$ and $+\infty$ by the following lemma, $\log p(\cdot|\mathbf{X})$ is always integrable with respect to $P$.
\begin{lemma}\label{lmm:Lambda_bounded_away}
    The value of $\Lambda_\epsilon(\beta, \mathbf{x})$ is bounded away from $0$ and $1$, for all $\beta\in\Beta$ and $\mathbf{x}\in\mathcal{X}$.
\end{lemma}
See \cref{app:proof_Lambda_bounded_away} for the proof.
Thus, if $F_\mathbf{X}\log F_\mathbf{X}$ exists, $P\log p$ also exists.
Similarly, the existence of $P\log p_\beta$ depends on the existence of $F_\mathbf{X}\log f_\mathbf{X}$.
\begin{assumption}\label{asm:FXlogfX_exists}
    $F_\mathbf{X}\log f_\mathbf{X}$ exists.
\end{assumption}

The second condition relates to the uniqueness of the maximum of the log-likelihood function.
Because the maxima are not always unique, we adopt the following assumption.
\begin{assumption}\label{asm:negative_definite}
    $P\log p_\beta$ has a unique maximum.
\end{assumption}

We now have consistency.
\begin{theorem}\label{thm:publicX_consistent}
    Suppose \cref{asm:fyx_measurable,asm:fbetayx_measurable,asm:FXlogfX_exists,asm:negative_definite,asm:fx_measurable} hold.
    Then, $\hat{\beta}_n\rightarrow \beta^*$ as $n\rightarrow\infty$ surely.
\end{theorem}

Next, we derive the asymptotic normality.
We find the conditions under which \cref{asm:A4,asm:A5,asm:A6} are satisfied.
\cref{asm:A4} specifies the continuous differentiability of $\partial \log p_\beta/\partial\beta$.
The partial derivative is extended as 
\[
    \frac{\partial}{\partial\beta}\log(p_{\beta}(\mathbf{x}, z))
    = \frac{(2z-1)\Lambda_\epsilon'(\beta,\mathbf{x})}{\Lambda_\epsilon(\beta,\mathbf{x})^{z}(1-\Lambda_\epsilon(\beta,\mathbf{x}))^{1-z}}
\]
where $\Lambda_\epsilon'(\beta, \mathbf{x}) = \partial \Lambda_\epsilon(\beta, \mathbf{x})/\partial \beta$.
By \cref{lmm:Lambda_bounded_away}, the following is sufficient to satisfy the requirement.
\begin{assumption}\label{asm:Lambda'_continuous}
    Each element of $\Lambda'_\epsilon(\beta, \mathbf{x})$ is measurable of $\mathbf{x}$ for each $\beta\in\Beta$ and continuously differentiable functions of $\beta$ for each $\mathbf{x}\in\mathcal{X}$.
\end{assumption}

\cref{asm:A5} states that $|\partial^2\log p_\beta/\partial\beta_{j_1}\partial\beta_{j_2}|$ and $|\partial\log p_\beta/\partial\beta_{j_1}\cdot\partial\log p_\beta/\partial\beta_{j_2}|$ for $j_1,j_2=1,\cdots, d$ are bounded by functions integrable with respect to $P$.
To verify this, we extend these values.
\begin{align*}
    \frac{\partial^2 \log p_\beta(\mathbf{x}, z)}{\partial \beta^2}
    = & (2z-1)\frac{\Lambda_\epsilon''(\beta, \mathbf{x})}{\Lambda_\epsilon(\beta, \mathbf{x})^{z}(1-\Lambda_\epsilon(\beta, \mathbf{x}))^{1-z}}\\
    & - \frac{\Lambda'_\epsilon(\beta, \mathbf{x})\Lambda'_\epsilon(\beta, \mathbf{x})^\top}{\Lambda_\epsilon(\beta, \mathbf{x})^{2z}(1-\Lambda_\epsilon(\beta, \mathbf{x}))^{2(1-z)}}
\end{align*}
where $\Lambda_\epsilon''(\beta, \mathbf{x})=\partial^2\Lambda_\epsilon(\beta, \mathbf{x})/\partial \beta^2$, and 
\begin{align*}
    & \left(\frac{\partial}{\partial \beta}\log p_\beta(\mathbf{x}, z)\right)\left(\frac{\partial}{\partial \beta}\log p_\beta(\mathbf{x}, z)\right)^\top\\
    = & \frac{\Lambda_\epsilon'(\beta, \mathbf{x})\Lambda_\epsilon'(\beta, \mathbf{x})^\top}{\Lambda_\epsilon(\beta, \mathbf{x})^{2z}(1-\Lambda_\epsilon(\beta,\mathbf{x}))^{2(1-z)}}.
\end{align*}
The denominators are always non-zero by \cref{lmm:Lambda_bounded_away}.
Thus, the following assumption is sufficient to satisfy the requirement.
\begin{assumption}\label{asm:Lambda'_bounded}
    The absolute values of each element of $\Lambda_\epsilon'(\beta, \mathbf{x})$ and $\Lambda_\epsilon''(\beta, \mathbf{x})$ are bounded by integrable functions with respect to $P$.
\end{assumption}

\cref{asm:A6} consists of three parts.
The first part is that $\beta^*$ is interior to $\Beta$.
We assume this.
\begin{assumption}\label{asm:beta_star_interior}
    $\beta^*$ is interior to $\Beta$.
\end{assumption}
The second part is the non-singularity of $P((\partial\log p_\beta/\partial\beta)(\partial\log p_\beta/\partial\beta)^\top)$ at $\beta=\beta^*$.
\begin{align*}
    & P\left(\frac{\partial}{\partial \beta}\log p_\beta\right)\left(\frac{\partial}{\partial \beta}\log p_\beta\right)^\top
    = \\
    & F_\mathbf{X}\left(\frac{p(Z=1|\mathbf{X})}{\Lambda_\epsilon(\beta, \mathbf{X})^2}+\frac{p(Z=0|\mathbf{X})}{(1-\Lambda_\epsilon(\beta, \mathbf{X}))^2}\right)
    \Lambda_\epsilon'(\beta, \mathbf{X})\Lambda_\epsilon'(\beta, \mathbf{X})^\top.
\end{align*}
Thus, the following assumption is a sufficient condition of the requirement.
\begin{assumption}\label{asm:X_spread}
    $F_\mathbf{X}\Lambda_\epsilon'(\beta^*, \mathbf{X})\Lambda_\epsilon'(\beta^*, \mathbf{X})^\top$ is non-singular.
\end{assumption}

The third part is non-singularity of $P\partial^2 \log p_\beta /\partial\beta^2$ at $\beta=\beta^*$.
We obtain this from \cref{asm:negative_definite}. 
If $P\log p_\beta$ has a second partial derivative along $\beta$ and $\beta^*$ is interior to $\Beta$, then $\partial^2P\log p_\beta/\partial\beta^2$ must be negative-definite.
If not, there exists $\beta'$ such that $P\log p_{\beta'} = P\log p_{\beta^*}$ and $\beta'\neq \beta^*$.

Finally, we obtain asymptotic normality.
\begin{theorem}\label{thm:public_normal}
    Suppose \cref{asm:fyx_measurable,asm:fbetayx_measurable,asm:FXlogfX_exists,asm:negative_definite,asm:X_spread,asm:beta_star_interior,asm:Lambda'_bounded,asm:fx_measurable,asm:Lambda'_continuous} hold.
    Then, $\sqrt{n}(\hat{\beta}_n-\beta^*) \to \mathcal{N}(0_d, C(\beta^*))$
    where $C(\beta) = A^{-1}(\beta)B(\beta)A^{-1}(\beta)$ with $A(\beta)=P\partial^2\log p_\beta/\partial\beta^2$ and $B(\beta)=P(\partial\log p_\beta/\partial\beta)(\partial\log p_\beta/\partial\beta)^\top$.
\end{theorem}

\subsection{Regression with Private $\mathbf{X}$}
\label{ssec:priate_X}

Next, we consider regression when both objective variables and explanatory variables are sensitive and are submitted with perturbation.
The system model is that each user $i$ generates $\mathbf{X}_i$ following unknown distribution $F_\mathbf{X}$ and then generates $Y_i$ following unknown conditional distribution $F(\cdot|\mathbf{X}_i)$.

The communication protocol is as follows.
User $i$ stochastically perturbs $\mathbf{X}_i$ and $Y_i$ by LDP mechanism $Q$.
We denote the perturbed ones by $\priX$ and $\priY$, respectively.
$Q$ consists of $Q_{\priY}$ and $Q_{\priX}$ perturbing $Y_i$ and $\mathbf{X}_i$, respectively.
The privatized objective variable $\priY$ is the same as $Z$ in the previous subsection without the privacy budget consumed by the LDP mechanisms.
On the other hand, since $\priX$ was not defined in the previous section, we need to define $Q_{\priX}$. 
We use the bit flip as $Q_{\priX}$ in an element-wise manner.
Each element is randomized with privacy budget $\epsilon/(k+1)$.
The total consumption of the privacy budget per user does not exceed $\epsilon$ by the sequential composition theorem~\citep{McSherry2009privacy}.
We set the domain of $Q_{\mathbf{Z}^{(\mathbf{X})}}$ to $\{-1, +1\}^k$.
For each $\mathbf{z}^{(\mathbf{X})}\in \{-1, +1\}^k$, 
\begin{align}
    Q_{\priX}(\prix|\mathbf{x})
    = \prod_{j=1}^k \left(\frac{1}{2}+\frac{t(x_j)z_j^{(\mathbf{X})}}{2C_{\epsilon/(k+1)}}\right).\label{eq:def_qzx}
\end{align}
The generated privatized variables $(\prixi, \priyi)$ are submitted to the curator.

In the communication protocol, each user submits $(k+1)$ bits to the curator, and the curator sends no information to the users.
This privacy scenario is nearly the same as the Bhowmich's one, and our communication protocol is more efficient than theirs.
In their protocol, each user receives and submits $d$ float or double values, either $64d$ bits or $144d$ bits.
Thus, our protocol results in communication costs that are roughly $64$ or $144$ times smaller than their protocol when $k\leq d$.

The curator defines model family $\{F_\beta(y|\mathbf{x}):\beta\in\Beta,\mathbf{x}\in\mathcal{X}\}$ and provisional distribution $\hat{F}_\mathbf{X}$.
Though the true $F_\mathbf{X}$ is unknown, the curator must assume some distribution of $\mathbf{X}$ to compute the log-likelihood function, as we will see later.
$\hat{F}_\mathbf{X}$ is a kind of prior distribution.

Since the discussion of consistency and asymptotic normality has much in common with the previous subsection, here we describe only the differences. 
See the appendix for details.
Given observations $D_n=\{(\prixi, \priyi)\}_{i=1}^n$, the likelihood function is 
\begin{align*}
    \nonumber
    \ell(\beta;D_n)
    \nonumber
    = & \frac{1}{n}\sum_{i=1}^n \big(\log \hat{p}_\priX(\prixi) + {z_i^{(Y)}}\log \Phi(\beta, \mathbf{z}_i^{(X)})\\
    & + {(1-z_i^{(Y)})}\log(1-\Phi(\beta, \mathbf{z}_i^{(\mathbf{X})}))\big)\\
    \text{where}\quad&\hat{p}_\priX(\prix) \equiv \int Q_{\priX}(\prix|\mathbf{x})d\hat{F}_\mathbf{X}(\mathbf{x}),\\
    & \Phi(\beta, \mathbf{z}^{(\mathbf{X})})
    \equiv \frac{\hat{F}_\mathbf{X}(\Lambda_{\epsilon/(d+1)}(\beta, \mathbf{X})Q_{\priX}(\priX|\mathbf{X})}{\hat{p}_\priX(\prix)}.
\end{align*}
QMLE $\hat{\beta}_n$ is defined as $\hat{\beta}_n\equiv \text{argmin}_{\beta\in\mathcal{B}}\ell(\beta;D_n)$.

We can show consistency based on \cref{thm:miss_consistent} under the assumption that the curator chooses a regular distribution as $\hat{F}_\mathbf{X}$.
\begin{assumption}\label{asm:hatf_measurable}
    $\hat{F}_\mathbf{X}$ has a measurable Radon-Nikodym density $\hat{f}_\mathbf{X}=d\hat{F}_\mathbf{X}/d\mu$. % with some appropriate measure $\mu$.
\end{assumption}
\begin{theorem}
    \label{thm:private_consistent}
    Suppose \cref{asm:fyx_measurable,asm:fbetayx_measurable,asm:hatf_measurable,asm:negative_definite,asm:fx_measurable} hold.
    Then, $\hat{\beta}_n\rightarrow \beta^*$ as $n\rightarrow\infty$ for almost every sequence $\{(\priXi, \priYi)\}_i$.
\end{theorem}
For details, see \cref{app:mathematical_private_X}.
This consistent theorem does not require the existence of $F_\mathbf{X}\log f_\mathbf{X}$ unlike \cref{thm:publicX_consistent}.
We can obtain the existence from \cref{asm:hatf_measurable} and the properties of $\hat{p}_{\priX}$.
The discretization by the bit flip relaxes the integrable condition.

To show asymptotic normality, we adopt several additional assumptions.
\begin{assumption}\label{asm:Phi_continuous_differentiable}
    $\Phi'(\beta, \prix)$ is continuous differentiable function of $\beta$.
\end{assumption}
\begin{assumption}\label{asm:Phi'_bounded}
    Each component of $\Phi''(\beta, \prix)$ and $(\Phi'(\beta, \prix))(\Phi'(\beta, \prix))^\top$ is bounded by integrable functions with respect to $P$.
\end{assumption}
\begin{assumption}\label{asm:Phi_regular}
    $\Expe{\priX}{(\Phi'(\beta, \priX)(\Phi'(\beta, \priX))^\top}$ is non-singular at $\beta=\beta^*$.
\end{assumption}
\cref{asm:Phi_continuous_differentiable} is used to prove the requiment corresponding to \cref{asm:A4}.

The requirement corresponding to \cref{asm:A5} is satisfied with \cref{asm:Phi'_bounded}, which requires that the curator should design $\Phi$ such that its first and second derivatives almost surely take finite values.
The requirement corresponding to \cref{asm:A6} is satisfied with \cref{asm:negative_definite,asm:beta_star_interior,asm:Phi_regular}.

We now have asymptotic normality.
\begin{theorem}
    \label{thm:normal_private}
    Suppose \cref{asm:fyx_measurable,asm:fbetayx_measurable,asm:hatf_measurable,asm:Phi_continuous_differentiable,asm:Phi_regular,asm:negative_definite,asm:Phi'_bounded,asm:beta_star_interior,asm:fx_measurable} hold.
    Then, $\sqrt{n}(\hat{\beta}_n-\beta^*) \to \mathcal{N}(0_d, C(\beta^*))$
    where $C(\beta) = A^{-1}(\beta)B(\beta)A^{-1}(\beta)$ with $A(\beta)=P\partial^2\log p_\beta/\partial\beta^2$ and $B(\beta)=P(\partial\log p_\beta/\partial\beta)(\partial\log p_\beta/\partial\beta)^\top$.
\end{theorem}

\subsection{Remark and Limitation}

The assumptions for proving consistency and asymptotic normality in \cref{thm:publicX_consistent,thm:public_normal,thm:private_consistent,thm:normal_private} are not relevant to privacy preservation.
Even if those assumptions do not hold, users' privacy is still protected as long as the $\epsilon$-LDP mechanisms correctly work.
The users who supply data do not need to worry about these assumptions at all.

The requirements of our theorems clarify the properties of the model that the curator should check.
The curator is free to choose any linear or non-linear model as long as it satisfies these properties. In addition, those requirements place few restrictions on model selection since the curator can modify the model after data collection. 

As we see in \cref{sec:example}, it is not so difficult to craft a model satisfying the requirements.
We thus expect that most standard regression models satisfy them.

The first limitation relates to the problem of choosing $\hat{F}$.
Although any $\hat{F}$ satisfying \cref{asm:hatf_measurable,asm:FXlogfX_exists} can be acceptable, a poor choice of $\hat{F}$ may make it difficult to satisfy the other assumptions.
The theorem provides no method for choosing a better $\hat{F}$, which remains an open problem.

The second limitation relates to the true distribution, which is a common problem in most statistical theories.
We have no method to evaluate \cref{asm:fyx_measurable,asm:negative_definite,asm:beta_star_interior}.
The curator never know the exact value of $\beta^*$ and $C(\beta^*)$.
The curator should carefully consider these assumptions with the help of experts.

The exploration of better mechanisms is our future work.
There may exist $Q_{\priX}$ giving us a more sharp covariance matrix.
In the context of LDP, vector submission is studied by many researchers e.g., \citep{Duchi2013minimax,Erlingsson2014rappor,Bassily2015succinct,Wang2019collecting}.

Better selection of $\lconst$ and $\uconst$ is another future work.
Whether certain $\lconst$ and $\uconst$ are good or bad strongly depends on $F$, and we have no general strategy to select better $\lconst$ and $\uconst$.

One of the potential applications of our algorithms is bootstrapping.
In the above subsections, we described that our algorithms output only one estimator in each protocol.
However, without additional privacy loss, the curator can compute many estimators using the subsets of the submitted data.
The post-processing invariant enables us to perform such an operation.
This is one of the advantages of a non-interactive algorithm.

Another potential application is a misspecification test to determin whether the model family contains the true distribution~\citep{White1982misspecified}.
In the LDP setting, since the raw data are distributed, no single entity has knowledge on the statistical properties of the raw data.
It is difficult to evaluate whether a model family is appropriate.
A curator performs the test as a preliminary experiment.
The results of the test would help the curator to quantitatively assess the confidence level of the main survey.

\section{EXAMPLE: QUANTILE REGRESSION}\label{sec:example}

In this section, we show the QMLEs for quantile regression as a concrete example of our QMLEs.
One of the main goals of this section is to show that it is possible to replace some of the assumptions noted in the previous section with a concrete implementation of the model. We note that the notation used in \cref{ssec:quantile_public} and \cref{ssec:quantile_private} is the same as that used in \cref{ssec:public_X} and \cref{ssec:priate_X}.
Here, $k=d$.

\subsection{With Public $\mathbf{X}$} \label{ssec:quantile_public}

As described in \Cref{ssec:quantile}, we can formulate the $\alpha$-quantile regression as a quasi-maximum likelihood estimation problem.
For some $\sigma>0$, we set $f_\beta$ as 
\begin{align*}
    f_\beta(y|\mathbf{x}) = 
    \frac{\alpha(1-\alpha)}{\sigma}\exp\left(-\rho_\alpha\left(\frac{y-\ip{\beta}{\mathbf{x}}}{\sigma}\right)\right)
\end{align*}
for each $y$ and $\mathbf{x}$, where $\rho_\alpha$ is defined in (\ref{eq:problem}).
This construction satisfies \cref{asm:fbetayx_measurable}: measurable and continuous.

When we choose the product of independent $d$ uniform distributions on interval $[-1, +1]$ as $F_\mathbf{X}$, \cref{asm:FXlogfX_exists} is satisfied.

Let $\Psi_\epsilon$ be the function such that $\Lambda_\epsilon(\beta, \mathbf{x}) = \Psi_\epsilon(\ip{\beta}{\mathbf{x}})$.
Then, $\Lambda'_\epsilon(\beta, \mathbf{x})=\Psi'_\epsilon(\ip{\beta}{\mathbf{x}})\mathbf{x}$ and $\Lambda''_\epsilon(\beta, \mathbf{x})=\Psi''_\epsilon(\ip{\beta}{\mathbf{x}})\mathbf{x}\mathbf{x}^\top$ where $\Psi'_\epsilon(\theta)=\partial\Psi_\epsilon(\theta)/\partial\theta$ and $\Psi''_\epsilon(\theta) = \partial^2\Psi_\epsilon(\theta)/\partial\theta^2$.
It has the following property.
\begin{lemma}\label{lmm:quantile_Psi}
    $\Psi_\epsilon(\theta)$ is a strictly monotonically increasing function and is bounded away from $0$ and $1$.
    $\Psi'_\epsilon(\theta)$ and $\Psi''_\epsilon(\theta)$ exist and for any $\theta\in\Real$, and their absolute values are bounded.
\end{lemma}
See \Cref{app:likelihood} for the proof.
From the second part of \cref{lmm:quantile_Psi}, \cref{asm:Lambda'_continuous,asm:Lambda'_bounded} are satisfied.
Mover, $F_\mathbf{X} (\mathbf{X}\mathbf{X}^\top)$ is a non-singular matrix since
\begin{align*}
    F_\mathbf{X} X_{j_1}X_{j_2} = 
    \begin{cases}
        0 & \text{if}~j_1\neq j_2,\\
        \frac{1}{3} & \text{if}~j_1=j_2.
    \end{cases}
\end{align*}
Thus, \Cref{asm:X_spread} is satisfied.

As a consequence of \Cref{thm:publicX_consistent,thm:normal_private}, we have the following corollaries.
\begin{corollary}\label{cor:quantile_normal_public}
    Suppose \cref{asm:fyx_measurable,asm:negative_definite,asm:beta_star_interior} hold.
    Then, $\hat{\beta}_n\to\beta^*$ almost surely and $\sqrt{n}(\hat{\beta}_n-\beta^*)\to \mathcal{N}(0_d, C(\beta^*))$.
\end{corollary}
To prove this corollary, we need only three assumptions.
The concrete constructions of the model remove some of the assumptions used in \Cref{thm:publicX_consistent,thm:normal_private}.

Although the accuracy of our QMLEs is not a focus of this paper, we did conduct a rough comparison of accuracy with existing works.
As a result, we found with $\epsilon\downarrow 0$, the Fisher information of our MLEs is $\sigma^2/\alpha(1-\alpha)$ times smaller than the upper bound shown in \citep{Barnes2020Fisher}.
For details, see \cref{app:compare}.

\subsection{With Private $\mathbf{X}$}
\label{ssec:quantile_private}

In this setting, the curator does not know $F_\mathbf{X}$.
Instead of $F_\mathbf{X}$, we adopt the product distribution of $d$ symmetric binary distributions on $\{-1, +1\}$.
Then, \cref{asm:hatf_measurable} is satisfied.

With $\epsilon'=\epsilon/(d+1)$, $\Phi$ is extended as 
\begin{align*}
    \Phi(\beta, \prix) = 
    \frac{\sum_{\mathbf{x}\in\{\pm 1\}^d}\Psi_{\epsilon'}(\ip{\beta}{\mathbf{x}})\exp\left(\epsilon'\indicator{\prixj=x_j}\right)}{p_{\priX}(\prix)(e^{\epsilon'}+1)^d2^d},
\end{align*}
where $\Psi_{\epsilon'}$ is defined in the previous subsection.
Due to the properties of $\Psi_{\epsilon'}$, which we evaluated in the previous subsection, \cref{asm:Phi_continuous_differentiable,asm:Phi'_bounded} are obviously satisfied.
By the monotonicity of $\Psi_{\epsilon'}$, \cref{asm:Phi_regular} is also satisfied.
Now, as a corollary of \cref{thm:private_consistent,thm:normal_private}, we obtain the following result.
\begin{corollary}
    Suppose \cref{asm:fyx_measurable,asm:negative_definite,asm:beta_star_interior} hold.
    Then, $\sqrt{n}(\hat{\beta}_n-\beta^*) \to \mathcal{N}(0_d, C(\beta^*))$ where $C(\beta) = A^{-1}(\beta)B(\beta)A^{-1}(\beta)$.
\end{corollary}

\section{NUMERICAL EVALUATION}\label{sec:numerical}

In this section, we observe the behavior of our QMLE for real data.
We consider the QMLE for quantile regression in the public $\mathbf{X}$ case.
Since we do not know the true distribution generating the real data, we cannot perform exact comparisons with the theoretical result, \cref{cor:normal_quantile_public}.
Here, we observe the empirical covariance of the QMLEs to evaluate the convergence of the distribution of the QMLE.
For additional numerical evaluations, see \cref{app:numerical}.

We numerically compare the covariance matrices with varying $n$ and $\epsilon$.
We use CO and NOx emission data set~\citep{Kaya2019nox}, which consists of $36,733$ records of $11$ sensors attached to a turbine of a power plant.
Although this data is not sensitive, we chose this data because of its large number of records and its format.
We treat the $11$th column as $y$ and treat the columns from the first to $9$th as $\mathbf{x}$.
We set $\uconst = 110, \lconst = 40, \sigma = 1.0$ and $\alpha = 0.3$.
These specific values of hyperparameters do not have a particular meaning.
We vary $n$ from $5,000$ to $35,000$ in increments of $5,000$ for $\epsilon\in\{1, 2.5, 5, 10\}$.
For each combination of $n$ and $\epsilon$, we sub-sample $n$ records $1,000$ times without replacement from the $36,733$ records.
For each sub-data, we perturb $y$s and compute a QMLE as descrived in \cref{ssec:quantile_public}.
With the $1,000$ QMLEs, we obtain the empirical covariance matrix and its Frobenius norm.
We implemented the simulations with Python 3.9.2, NumPy 1.19.2, and SciPy 1.6.1.
The Python code is contained in the supplementary material.

\Cref{fig:covs_emission} shows the result.
The horizontal and vertical axes show $n$ and the value of each Frobenius norm in log-scale, respectively.
For each $\epsilon$, with large $n$, the norm of the covariance matrix is smaller.
The decreasing speed is $O(1/n)$, and this result is compatible with the theoretical result.
Greater $\epsilon$ also gives smaller covariance.
In this case, the QMLE is concentrated in one point, and, as $n$ increases, the distribution becomes more concentrated at that point.

\begin{figure}[t]
    \centering
    \includegraphics[scale=0.45]{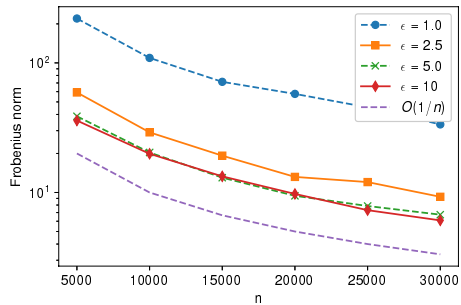}
    \caption{Frobenius norm of covariance matrices. The norms decrease in proportion to $1/n$ for each $\epsilon$. }\label{fig:covs_emission}
\end{figure}

\section{CONCLUSION}\label{sec:conclusion}

We developed the simple protocols for building QMLEs from distributed data while guaranteeing $\epsilon$-LDP for the users.
They address the two different privacy scenarios.
In the protocols, users submit only one or a few bits to the curator and do not need to wait for one another.
Moreover, the users do not need to perform complex computations such as integration or derivation.
Thus, the protocols are highly user-friendly and suitable for low-priced devices.
We clarified the sufficient conditions for the QMLEs to be consistent and asymptotically normal, and showed their limitations.
We showed that the sufficient conditions are relaxed with a concrete implementation.

\subsubsection*{Acknowledgements}
H.H. is partly supported by JSPS KAKENHI No. JP19H04113 and CREST JST CREST Grant No. JPMJCR2015, and K.M. is partly supported by JSPS KAKENHI No.JP21H04403.

\bibliographystyle{unsrtnat}
\bibliography{ref}

%%%%%%%%%%%%%%%%%%%%%%%%%%%%%%%%%%%
%%%%%% SUPPLEMENT (OPTIONAL) %%%%%%
%%%%%%%%%%%%%%%%%%%%%%%%%%%%%%%%%%%

\clearpage
\appendix

\thispagestyle{empty}

% For one-column format, uncomment the following:
%\onecolumn \makesupplementtitle
% For two-column format, uncomment the following:
%\twocolumn[ \makesupplementtitle ]

\section{SUMMARY OF NOTATION}\label{app:notation}

\subsection{Defined in \cref{sec:preliminaries}}

$\bitf$ is the bit flip, and $C_\epsilon=\frac{e^\epsilon+1}{e^\epsilon-1}$ is a value used to define the bit flip.

With $0<\alpha<1$, check loss for $\alpha$-quantile is defined as
\begin{align*}
& \checkl_\alpha(\tau) = 
  \begin{cases}
    (\alpha - 1) \tau & \text{if}\hspace{2pt} \tau \leq 0,\\
    \alpha \tau & \text{if}\hspace{2pt} \tau > 0.
  \end{cases}
\end{align*}

\subsection{Defined in \cref{ssec:public_X}}

$Y$ and $\mathbf{X}$ are objective and explanatory variables.
$F_\mathbf{X}$ is the distribution of $\mathbf{X}$, and $F(\cdot|\mathbf{X}=\mathbf{x})$ is the distribution of $Y$ conditioned by $\mathbf{X}=\mathbf{x}$.
$\mathcal{X}$ and $\mathcal{Y}$ are the domains of $\mathbf{X}$ and $\mathbf{Y}$.
For each $i=1,\cdots,n$, $(\mathbf{X}_i, Y_i)$ is an independent copy of $(\mathbf{X}, Y)$, which is possessed by the $i$th user.
$\bar{Y}_i$ is truncated version of $Y_i$, and $t(\cdot)$ is the truncating function mapping $\mathcal{Y}$ into $[\lconst, \uconst]$, where $\lconst$ and $\uconst$ are real values such that $\lconst<\uconst$.
$Z_i$ is the perturbed version of $Y_i$, and its distribution is 
\begin{align*}
    p(Z_i=z|\mathbf{X}_i=\mathbf{x})
    = \int \bitf(z|t(y)) dF(y|\mathbf{x}).
\end{align*}

$\{F_\beta(\cdot|\mathbf{x}):\beta\in\Beta, \mathbf{x}\in\mathcal{X}\}$ is the model family and  $\Beta$ is the parameter set.
For each $\beta\in\Beta$, $p_\beta(z|\mathbf{x})$ is the density function which is obtained by replacing $F$ by $F_\beta$ in (\ref{eq:p_public}).
In this \cref{ssec:public_X}, we write $P$ and $P_\beta$ to designate joint distributions $P(\mathbf{x},z)$ and $P_\beta(\mathbf{x},z)$ rather than conditional distributions $P(z|\mathbf{x})$ and $P_\beta(z|\mathbf{x})$.

$\Lambda_\epsilon(\beta, \mathbf{x}) = p_\beta(z=1|\mathbf{x})$.
Its first and second derivatives along $\beta$ are denoted by $\Lambda'_\epsilon(\beta, \mathbf{x})$ and $\Lambda''_\epsilon(\beta, \mathbf{x})$.

\subsection{Defined in \cref{ssec:priate_X}}

$F_\mathbf{X}$ is the model of $\mathbf{X}$.
$Q$ consists of $Q_{\priY}$ and $Q_{\priX}$ perturbing $Y_i$ and $\mathbf{X}_i$, respectively.

$\hat{p}(\prix)$ is the model of $\prix$ written as
\begin{align*}
     \hat{p}_X(\prix) = \int Q_{\priX}(\prix|\mathbf{x})d\hat{F}(\mathbf{x}).
\end{align*}

$\Phi(\beta, \mathbf{z}^{(\mathbf{X})})$ is the probability that $\priY|\mathbf{X}=\mathbf{x}$ is $1$ when model $F_\beta$ is correct:
\begin{align*}
    \Phi(\beta, \prix)
    = \frac{\hat{F}_\mathbf{X}(\Lambda_{\epsilon/(d+1)}(\beta, \mathbf{X})Q_{\priX}(\priX|\mathbf{X})}{\hat{p}_\priX(\prix)}.
\end{align*}
Its first and second derivatives along $\beta$ are denoted by $\Phi'(\beta, \prix)$ and $\Phi''(\beta, \prix)$.

\section{PSEUDO-CODE}\label{app:pseudo}

\cref{alg:public} and \cref{alg:private} are the pseudo-codes of the protocols described in \cref{ssec:public_X} and \cref{ssec:priate_X}, respectively.
In the for loops, the processing of each user does not need to be synchronized.

\begin{algorithm}[t]
    \SetKwComment{Comment}{/* }{ */}
    \caption{Protocol with Public $\mathbf{X}$}\label{alg:public}
    \KwIn{Unknown distribution $F$, privacy parameter $\epsilon$ and $\Beta$}
    Curator set $F_X$\;
    \For{$i=1$ to $n$}{
        Curator generates $\mathbf{x}_i\sim F_X$\;
        Send $\mathbf{x}_i$ to user $i$\;
        User $i$ generates $y_i\sim F(\cdot|\mathbf{x}_i)$\;
        User $i$ computes $\bar{y}_i$ as (\ref{eq:truncated})\;
        User $i$ generates $z_i\sim \bitf(\cdot|\bar{y}_i)$\;
        Send $z_i$ to curator\;
    }
    Let $D_n=\{(\mathbf{x}_i, z_i)\}_{i=1 }^n$\;
    Curator computes $\ell(\beta;D)$\;
    Computes $\hat{\beta}_n=\argmax_{\beta\in\Beta}\ell(\beta;D)$\;
    \KwOut{$\hat{\beta}_n$}
\end{algorithm}
\begin{algorithm}[t]
    \caption{Protocol with Private $\mathbf{X}$}\label{alg:private}
    \KwIn{Unknown distribution $F, F_X$, privacy parameter $\epsilon$ and $\Beta$}
    Curator set $\hat{F}_X$\;
    \For{$i=1$ to $n$}{
        User $i$ generates $\mathbf{x}_i\sim F_X$\;
        User $i$ generates $y_i\sim F(\cdot|\mathbf{x}_i)$\;
        User $i$ computes $\bar{y}_i$ as (\ref{eq:truncated})\;
        User $i$ generates $\priyi\sim \bitf(\cdot|\bar{y}_i)$\;
        \For{$j=1$ to $d$}{
            User $i$ compute $\bar{x}_{ij}$ as (\ref{eq:truncated})\;
            Generate $\prixij\sim\bitf(\cdot|\bar{x}_{ij})$\;
        }
        Let $\prixi=(\prixij)_i$\;
        Send $(\prixi, \priyi)$ to curator\;
    }
    Let $D_n=\{(\prixi, \priyi)\}_{i=1 }^n$\;
    Curator computes $\ell(\beta;D)$\;
    Computes $\hat{\beta}_n=\argmax_{\beta\in\Beta}\ell(\beta;D)$\;
    \KwOut{$\hat{\beta}_n$}
\end{algorithm}

\section{MATHEMATICAL NOTES}\label{app:mathematical_notes}

\subsection{for \cref{ssec:public_X}}

\subsubsection{Proof of \cref{lmm:Lambda_bounded_away}}\label{app:proof_Lambda_bounded_away}

By the definition of $\Lambda_\epsilon(\beta, \mathbf{x})$, it is written as 
\begin{align*}
    \Lambda_\epsilon(\beta, \mathbf{x})
    = p_\beta(Z=1|\mathbf{X}=\mathbf{x})
    = \int \bitf(1|t(y))dF_\beta(y|\mathbf{x}).
\end{align*}
From the definition of $\bitf$, we have
\begin{align*}
    \frac{1}{e^\epsilon+1} \leq \bitf(1|t(y)) \leq \frac{e^\epsilon}{e^\epsilon+1}
\end{align*}
for any $y\in\mathcal{Y}$.
Thus, the following relation holds.
\begin{align*}
    \Lambda_\epsilon(\beta, \mathbf{x})
    \leq \int \frac{e^\epsilon}{e^\epsilon+1}dF_\beta(y|\mathbf{x})
    = \frac{e^\epsilon}{e^\epsilon+1}.
\end{align*}
The last equation is by the fact that $F_\beta$ is a probability distribution.
Similarly, we have
\begin{align*}
    \Lambda_\epsilon(\beta, \mathbf{x})\geq \frac{1}{e^\epsilon+1}.
\end{align*}

\subsection{for \cref{ssec:priate_X}}
\label{app:mathematical_private_X}

For each $\prix\in\{-1, +1\}^d$, the curator considers the probability distribution of $\priX$ at $\prix$ as 
\begin{align*}
    \hat{p}_\priX(\prix) = \int Q_{\priX}(\prix|\mathbf{x})d\hat{F}(\mathbf{x}).
\end{align*}
The joint density is 
\begin{align*}
    p_\beta(\prix, \priy)
    = \int Q_{\priX}(\prix|\mathbf{x})Q_{\priY}(\priy|t(y))dF_\beta(y|\mathbf{x})d\hat{F}_X(\mathbf{x}).
\end{align*}

The conditional distribution of $\priY$ is written as 
\begin{align*}
    p_\beta(\mathbf{z}^{(Y)}|\mathbf{z}^{(\mathbf{X})})
    = \frac{\hat{F}_X(p_\beta(\priy|\mathbf{X})Q_{\mathbf{Z}^{(\mathbf{X})}}(\mathbf{z}^{(\mathbf{X})}|\mathbf{X}))}{\hat{p}_\priX(\prix)}
    = \Phi(\beta, \mathbf{z}^{(\mathbf{X})})^{z^{(Y)}}
    (1-\Phi(\beta, \mathbf{z}^{(\mathbf{X})}))^{1-z^{(Y)}}.
\end{align*}

With $\Phi$, the joint density is written as 
\begin{align*}
    p_\beta(\prix, \priy)
    = \Phi(\beta, \mathbf{z}^{(\mathbf{X})})^{z^{(Y)}}
    (1-\Phi(\beta, \mathbf{z}^{(\mathbf{X})}))^{1-z^{(Y)}}\hat{p}_\priX(\prix).
\end{align*}

We analyze the sufficient conditions under which \cref{asm:A1,asm:A2,asm:A3} are satisfied while replacing $F$ and $F_\theta$ in \cref{thm:miss_consistent} with $P$ and $P_\beta$.
We adopt \cref{asm:fyx_measurable,asm:fbetayx_measurable,asm:fx_measurable,asm:hatf_measurable}.
From these assumptions, it is obvious that $P(\prix, \priy)$ and $P_\beta(\prix, \priy)$ are measurable, and that density functions $p(\prix, \priy)$ and $p_\beta(\prix, \priy)$ exist.

The condition corresponding to \cref{asm:A3} consists of two parts.
The first part is the existence of $P\log p$
integrable function $h(\prix, \priy)$ such that $|\log p_\beta(\prix, \priy)|\leq h(\prix, \priy)$ for all $\beta$.
$P\log p$ is expanded as follows:
\[
    P\log p = P_\priX (P_{\cdot|\priX} \log p(\cdot|\priX) + \log p_\priX).
\]
To evaluate the bound condition, it is necessary to analyze $p_{\priX}$ and $p(\cdot|\priX)$.
\begin{lemma}\label{lmm:hatpx_bounded}
    For any $\prix\in\{-1, +1\}^d$, 
    \[
        \left(\frac{1}{e^{\epsilon/(d+1)}+1}\right)^d
        \leq 
        \hat{p}_\priX(\prix)
        \leq 
        \left(\frac{e^{\epsilon/(d+1)}}{e^{\epsilon/(d+1)}+1}\right)^d.
    \]
    $p_\priX(\prix)$ has the same bounds.
\end{lemma}
\begin{lemma}\label{lmm:Phi_bounded_away}
    For any $\beta\in\Beta$ and $\prix\in\{-1, +1\}^d$, 
    \begin{equation*}
        \frac{1}{e^{\epsilon/(d+1)}+1}
        \leq \Phi(\beta, \mathbf{z}^{(\mathbf{X})})
        \leq 
        \frac{e^{\epsilon/(d+1)}}{e^{\epsilon/(d+1)}+1}.
    \end{equation*}
    With $0<\epsilon<+\infty$ and $1\leq d< +\infty$, $\Phi(\beta, \mathbf{z}^{(\mathbf{X})})$ are always bounded away from $0$ and $1$.
    Also, 
    \begin{equation*}
        \frac{1}{e^{\epsilon/(d+1)}+1}
        \leq 1-\Phi(\beta, \mathbf{z}^{(\mathbf{X})})
        \leq 
        \frac{e^{\epsilon/(d+1 )}}{e^{\epsilon/(d+1)}+1}.
    \end{equation*}
\end{lemma}
By the above lemmas, $\log p(\cdot|\prix)$ and $\log p_\priX$ are always bounded away from $\pm\infty$, and $\log p(\cdot|\prix)$ and $\log p_\priX$ are always integrable with respect to $P$.
The existence of integral function $h$ is also obtained.

The second part is the uniqueness of the log-likelihood function.
To guarantee that this property holds, we again adopt \cref{asm:negative_definite}.
Then, we have \cref{thm:private_consistent}.

We next analyze the conditions under which \cref{asm:A4,asm:A5,asm:A6} are satisfied.
The condition corresponding to \cref{asm:A4} is the continuous differentiability of $\partial \log p_\beta/\partial \beta$.
The partial derivative is expanded as 
\begin{align*}
    \frac{\partial}{\partial \beta} \log p_{\beta}(\prix, \priy)
    =& \priy \frac{\Phi'(\beta, \prix)}{\Phi(\beta, \prix)} - (1-\priy)\frac{\Phi'(\beta, \prix)}{1-\Phi(\beta, \prix)}\\
    =& \frac{\Phi'(\beta,\prix)(\priy-\Phi(\beta, \prix))}{\Phi(\beta, \prix)(1-\Phi(\beta, \prix))}
\end{align*}
where
\[
    \Phi'(\beta, \prix) \equiv \frac{\partial}{\partial \beta} \Phi(\beta, \prix).
\]
By \cref{lmm:Phi_bounded_away}, $\Phi(\beta, \mathbf{z}^{(\mathbf{X})})$ always takes values greater than $0$ and less than $1$.
So, if \cref{asm:Phi_continuous_differentiable} holds, \cref{asm:A4} is satisfied.

The condition corresponding to \cref{asm:A5} is that there exist integrable functions with respect to $P$ that upper bound the absolute values of each component of $\partial^2\log p_\beta(\prix, \priy)/\partial\beta^2$ and $(\partial\log p_\beta(\prix, \priy)/\partial\beta)(\partial\log p_\beta(\prix, \priy)/\partial\beta)^\top$.
The second-order derivative is 
\begin{align*}
    \frac{\partial^2}{\partial \beta^2} \log p_{\beta}(\prix, \priy)
    =& (2\priy-1)\frac{\Phi''(\beta, \prix)}{\Psi_\epsilon(\ip{\beta}{x})^{z}(1-\Psi_\epsilon(\ip{\beta}{x}))^{1-\priy}}\\
    &- \frac{\Phi'(\beta, \prix))(\Phi'(\beta, \prix))^\top}{\Phi(\beta, \prix)^{2\priy}(1-\Phi(\beta, \prix))^{2(1-\priy)}}
\end{align*}
where we define $\Phi''(\beta, \prix) \equiv \frac{\partial^2}{\partial \beta^2}\Phi(\beta, \prix)$.
$(\partial\log p_\beta(\prix, \priy)/\partial\beta)(\partial\log p_\beta(\prix, \priy)/\partial\beta)^\top$ is 
\begin{align*}
    &\left(\frac{\partial}{\partial \beta} \log p_{\beta}(\prix, \priy)\right)
    \left(\frac{\partial}{\partial \beta} \log p_{\beta}(\prix, \priy)\right)^\top\\
    &= \left(\frac{z-\Phi(\beta, \prix)}{\Phi(\beta, \prix)(1-\Phi(\beta, \prix))}\right)^2\Phi'(\beta,\prix)\Phi'(\beta,\prix)^{\top}
\end{align*}
By \cref{lmm:Phi_bounded_away}, \cref{asm:Phi'_bounded} is sufficient to make the requirement hold.

The requirement corresponding \cref{asm:A6} consists of three parts.
The first part is that $\beta$ is interior of $\Beta$.
We assume this as \cref{asm:beta_star_interior}.
For enough large $\Beta$, this assumption is not particularly strong.
Letting
\begin{align*}
    A(\beta) = & P \frac{\partial^2}{\partial\beta^2}\log p_\beta\quad\text{and}\quad
    B(\beta) = P \left(\frac{\partial}{\partial\beta}\log p_\beta\right)\left(\frac{\partial}{\partial\beta}\log p_\beta\right)^\top, 
\end{align*}
the second and third parts are the regularity of $A(\beta^*)$ and $B(\beta^*)$.
We have already assumed that $A(\beta^*)$ is regular in \cref{asm:negative_definite}.
We consider the regularity of $B(\beta^*)$ here.
$B(\beta)$ is 
\begin{align*}
    B(\beta)
    = 
    \mathbb{E}_{\priX}\Bigg[&\left(\frac{p(1|\priX)}{\Phi(\beta, \priX)^2}+\frac{p(0|\priX)}{(1-\Phi(\beta, \priX))^2}\right)(\Phi'(\beta, \priX)(\Phi'(\beta, \priX))^\top\Bigg]
\end{align*}
Since the scalar part is always finite and positive, \cref{asm:Phi_regular} is a sufficient condition of the regularity of $B(\beta^*)$.

Summarizing the above discussions, we obtain \cref{thm:normal_private}.

\subsubsection{Proof of \cref{lmm:hatpx_bounded}}

\begin{proof}
    By definition, for any $\prix\in\{-1, +1\}^d$, we have
    \begin{align*}
        \hat{p}_X(\prix) =  \int Q_{\priX}(\prix|\mathbf{x})d\hat{F}(\mathbf{x})
        \leq  \int \left(\frac{e^{\epsilon/(d+1)}}{e^{\epsilon/(d+1)}+1}\right)^dd\hat{F}(\mathbf{x})
        =  \left(\frac{e^{\epsilon/(d+1)}}{e^{\epsilon/(d+1)}+1}\right)^d.
    \end{align*}
    Similarly, we have
    \begin{align*}
        \hat{p}_X(\prix)
        \geq \left(\frac{1}{e^{\epsilon/(d+1)}+1}\right)^d.
    \end{align*}
\end{proof}

\subsubsection{Proof of \cref{lmm:Phi_bounded_away}}

\begin{proof}
    By \cref{lmm:Lambda_bounded_away} and (\ref{eq:def_qzx}), we have
    \begin{align*}
        \Phi(\beta, \mathbf{z}^{(\mathbf{X})})
        = & \frac{\hat{F}_X{\Lambda(\beta,\mathbf{X})Q_{\mathbf{Z}^{(\mathbf{X})}}(\mathbf{z}^{(\mathbf{X})}|\mathbf{X})}}{p_\priX(\prix)}
        \leq \frac{\hat{F}_X{\frac{e^{\epsilon/(d+1 )}}{e^{\epsilon/(d+1)}+1}Q_{\mathbf{Z}^{(\mathbf{X})}}(\mathbf{z}^{(\mathbf{X})}|\mathbf{X})}}{p_\priX(\prix)}
        = \frac{e^{\epsilon/(d+1)}}{e^{\epsilon/(d+1)}+1}.
    \end{align*}
    Similarly, we have
    \[
        \Phi(\beta, \mathbf{z}^{(\mathbf{X})})
        \geq \frac{1}{e^{\epsilon/(d+1)}+1}.
    \]
    Replacing $\hat{F}_X$ by $F_X$, we obtain the arguments with regard to $p(\priy|\prix)$.
\end{proof}

\subsection{for \cref{sec:example}}\label{app:likelihood}

In this section, we derive $\Psi_{\epsilon}(\theta)$ used in \Cref{sec:example}.
As a consequence of the analysis, we obtain \cref{lmm:quantile_Psi}.
We  analyze the function in different three cases.
The first case is the case where $\lconst<\theta<\uconst$.
For the sake of simplicity of notation, we let $G=\exp\left(-\frac{\alpha-1}{\sigma}(\lconst-\theta)\right)$ and $H=\exp\left(-\frac{\alpha}{\sigma}(\uconst-\theta)\right)$.
These values appear many times throughout the remainder of this section.
First, we extend the probability $F_\theta(Y_i\leq \lconst)$.
\begin{align*}
    F_\theta(Y_i\leq \lconst)
    = & \int_{-\infty}^{\lconst}\frac{\alpha(1-\alpha)}{\sigma}\exp\left(-\rho\left(\frac{y_i-\theta}{\sigma}\right)\right)dy_i\\
    = & \int_{-\infty}^{\lconst}\frac{\alpha(1-\alpha)}{\sigma}\exp\left(-\frac{\alpha-1}{\sigma}(y_i-\theta)\right)dy_i\\
    = & \left[\frac{\alpha(1-\alpha)}{\sigma}\left(-\frac{\sigma}{\alpha-1}\right)\exp\left(-\frac{\alpha-1}{\sigma}(y_i-\theta)\right) \right]_{-\infty}^{\lconst}\\
    = & \alpha \exp\left(-\frac{\alpha-1}{\sigma}(\lconst-\theta)\right) - \alpha \times 0\\
    = & \alpha \exp\left(-\frac{\alpha-1}{\sigma}(\lconst-\theta)\right)=\alpha G.
\end{align*}
Similarly, the probability $F_\theta(Y_i\geq \uconst)$ is expanded as:
\begin{align*}
    F_\theta(Y_i\geq \uconst)
    = & \int^{+\infty}_{\uconst}\frac{\alpha(1-\alpha)}{\sigma}\exp\left(-\rho\left(\frac{y_i-\theta}{\sigma}\right)\right)dy_i\\
    = & \int^{+\infty}_{\uconst}\frac{\alpha(1-\alpha)}{\sigma}\exp\left(-\frac{\alpha}{\sigma}(y_i-\theta)\right)dy_i\\
    = & \left[\frac{\alpha(1-\alpha)}{\sigma}\left(-\frac{\alpha}{\sigma}\right)\exp\left(-\frac{\alpha}{\sigma}(y_i-\theta)\right)\right]^{+\infty}_{\uconst}\\
    = & -(1-\alpha)\times 0 + (1-\alpha)\exp\left(-\frac{\alpha}{\sigma}(\uconst-\theta)\right)\\
    = & (1-\alpha)\exp\left(-\frac{\alpha}{\sigma}(\uconst-\theta)\right)=(1-\alpha)H.
\end{align*}
The probability $P_\theta(Z_i=1)$ is written as follows:
\begin{align}
    \nonumber
    P_\theta(Z_i=1)
    = & \alpha G\left(\frac{1}{2}-\frac{1}{2C_\epsilon}\right)\\
    \nonumber
    & + \frac{\alpha(1-\alpha)}{\sigma}\int_{\lconst}^{\theta}\left(\frac{1}{2}+\frac{y_i-\frac{\uconst+\lconst}{2}}{C_\epsilon(\uconst-\lconst)}\right)\exp\left(-(\alpha-1)\frac{y_i-\theta}{\sigma}\right)dy_i\\
    \nonumber
    & + \frac{\alpha(1-\alpha)}{\sigma}\int_{\theta}^{\uconst}\left(\frac{1}{2}+\frac{y_i-\frac{\uconst+\lconst}{2}}{C_\epsilon(\uconst-\lconst)}\right)\exp\left(-\alpha\frac{y_i-\theta}{\sigma}\right)dy_i\\
    & + (1-\alpha)H\left(\frac{1}{2}+\frac{1}{2C_\epsilon}\right)\label{eq:prob_Z_1}
\end{align}
Now, we extend each term.
\begin{align}
    \nonumber
    & \frac{\alpha(1-\alpha)}{\sigma}\int_{\lconst}^{\theta}\left(\frac{1}{2}+\frac{y_i-\frac{\uconst+\lconst}{2}}{C_\epsilon(\uconst-\lconst)}\right)\exp\left(-(\alpha-1)\frac{y_i-\theta}{\sigma}\right)dy_i\\
    \nonumber
    = & \frac{\alpha(1-\alpha)}{\sigma}\left(\frac{1}{2}-\frac{\uconst+\lconst}{2C_\epsilon(\uconst-\lconst)}+\frac{\theta}{C_\epsilon(\uconst-\lconst)}\right)\int_{\lconst}^{\theta}\exp\left(-(\alpha-1)\frac{y_i-\theta}{\sigma}\right)dy_i\\
    \nonumber
    & + \frac{\alpha(1-\alpha)}{\sigma}\frac{1}{C_\epsilon(\uconst-\lconst)}\int_{\lconst}^{\theta}(y_i-\theta)\exp\left(-(\alpha-1)\frac{y_i-\theta}{\sigma}\right)dy_i\\
    \nonumber
    = & \frac{\alpha(1-\alpha)}{\sigma}\left(\frac{1}{2}-\frac{\uconst+\lconst}{2C_\epsilon(\uconst-\lconst)}+\frac{\theta}{C_\epsilon(\uconst-\lconst)}\right)\frac{\sigma}{1-\alpha}(1-G)\\
    \nonumber
    & + \frac{\alpha(1-\alpha)}{2\sigma}\frac{1}{C_\epsilon(\uconst-\lconst)}\bigg(-\frac{\sigma^2}{(1-\alpha)^2}-\frac{\sigma}{1-\alpha}(\lconst-\theta)G
    +\frac{\sigma^2}{(1-\alpha)^2}G\bigg)\\
    = & \alpha\left(\frac{1}{2}-\frac{\uconst+\lconst}{2C_\epsilon(\uconst-\lconst)}+\frac{\theta}{C_\epsilon(\uconst-\lconst)}\right)(1-G)
    +\frac{\alpha}{C_\epsilon(\uconst-\lconst)} \bigg(-\frac{\sigma}{1-\alpha}-(\lconst-\theta)G+\frac{\sigma}{1-\alpha}G\bigg).\label{eq:part1}
\end{align}
Similarly, 
\begin{align}
    \nonumber
    & \frac{\alpha(1-\alpha)}{\sigma}\int_{\theta}^{\uconst}\left(\frac{1}{2}+\frac{y_i-\frac{\uconst+\lconst}{2}}{C_\epsilon(\uconst-\lconst)}\right)\exp\left(-\alpha\frac{y_i-\theta}{\sigma}\right)dy_i\\
    \nonumber
    = & -(1-\alpha)\left(\frac{1}{2}-\frac{\uconst+\lconst}{2C_\epsilon(\uconst-\lconst)}+\frac{\theta}{C_\epsilon(\uconst-\lconst)}\right)(H-1)\\
    &+ \frac{\alpha(1-\alpha)}{\sigma C_\epsilon(\uconst-\lconst)}\bigg(-\frac{\sigma}{\alpha}(\uconst-\theta)H-\frac{\sigma^2}{\alpha^2}H + \frac{\sigma^2}{\alpha^2}\bigg)\\
    \nonumber
    = & -(1-\alpha)\left(\frac{1}{2}-\frac{\uconst+\lconst}{2C_\epsilon(\uconst-\lconst)}+\frac{\theta}{C_\epsilon(\uconst-\lconst)}\right)(H-1)\\
    & + \frac{1-\alpha}{C_\epsilon(\uconst-\lconst)}\bigg(-(\uconst-\theta)H-\frac{\sigma}{\alpha}H+\frac{\sigma}{\alpha}\bigg).\label{eq:part2}
\end{align}
Substituting (\ref{eq:part1}) and (\ref{eq:part2}) into (\ref{eq:prob_Z_1}), we have
\begin{align*}
    & \Psi_{\epsilon}(\theta) = P_\theta(Z_i=1)\\
    = & \frac{\theta}{C_\epsilon(\uconst-\lconst)}+\frac{\alpha}{1-\alpha}\frac{\sigma}{C_\epsilon(\uconst-\lconst)}\exp\left(-\frac{\alpha-1}{\sigma}(\lconst-\theta)\right) - \frac{1-\alpha}{\alpha}\frac{\sigma}{C_\epsilon(\uconst-\lconst)}\exp\left(-\frac{\alpha}{\sigma}(\uconst-\theta)\right)\\
    & + \frac{1}{2} + \left(-\frac{\alpha}{1-\alpha}+\frac{1-\alpha}{\alpha}\right)\frac{\sigma}{C_\epsilon(\uconst-\lconst)}- \frac{\uconst+\lconst}{2C_\epsilon(\uconst-\lconst)}.
\end{align*}
The first and second derivatives are
\begin{align*}
    \Psi_{\epsilon}'(\theta) = &  \frac{1}{C_\epsilon(\uconst-\lconst)} - \frac{\alpha}{C_\epsilon(\uconst-\lconst)}\exp\left(\frac{1-\alpha}{\sigma}(\lconst-\theta)\right)
    - \frac{1-\alpha}{C_\epsilon(\uconst-\lconst)}\exp\left(-\frac{\alpha}{\sigma}(\uconst-\theta)\right),\\
    \Psi''_\epsilon(\theta)
    = & \frac{\alpha(1-\alpha)}{\sigma C_\epsilon(\uconst-\lconst)}
    \Bigg(e^{\frac{1-\alpha}{\sigma}(\lconst-\theta)}
    - e^{-\frac{\alpha}{\sigma}(\uconst-\theta)}\Bigg).
\end{align*}
By $\lconst < \theta < \uconst$, $\frac{1-\alpha}{\sigma}(\lconst-\ip{\beta}{\mathbf{x}})$ and $-\frac{\alpha}{\sigma}(\uconst-\ip{\beta}{\mathbf{x}})$ are always negative.
\begin{align*}
    |\Psi_\epsilon'(\theta)|<\frac{1}{C_\epsilon(\uconst-\lconst)}
    \quad \text{and}\quad
    |\Psi_\epsilon''(\theta)|<\frac{\alpha(1-\alpha)}{\sigma C_\epsilon(\uconst-\lconst)}.
\end{align*}

The second case is the case where $\theta\leq\lconst$.
$\Psi_\epsilon(\theta)$ is computed as 
\begin{align*}
    \Psi_\epsilon(\theta) = & \left(-(1-\alpha)\exp\left(-\alpha\frac{\lconst-\theta}{\sigma}\right)+1\right)\left(\frac{1}{2}-\frac{1}{2C_\epsilon}\right)\\
    & + \frac{\alpha(1-\alpha)}{\sigma}\frac{1}{C_\epsilon(\uconst-\lconst)}\left(\frac{\sigma}{\alpha}(\lconst-\theta)+\frac{\sigma^2}{\alpha^2}\right)\exp\left(-\frac{\alpha}{\sigma}(\lconst-\theta)\right)\\
    & - \frac{\alpha(1-\alpha)}{\sigma}\frac{1}{C_\epsilon(\uconst-\lconst)}\left(\frac{\sigma}{\alpha}(\uconst-\theta)+\frac{\sigma^2}{\alpha^2}\right)\exp\left(-\frac{\alpha}{\sigma}(\uconst-\theta)\right)\\
    & +\frac{\alpha(1-\alpha)}{\sigma}\left(\frac{1}{2}+\frac{\theta}{C_\epsilon(\uconst-\lconst)}-\frac{\uconst+\lconst}{2C_\epsilon(\uconst-\lconst)}\right)\frac{\sigma}{\alpha}
    (\exp(-\frac{\alpha}{\sigma}(\lconst-\theta))-\exp(-\frac{\alpha}{\sigma}(\uconst-\theta)))\\
    & + (1-\alpha)\exp\left(-\frac{\alpha}{\sigma}(\uconst-\theta)\right)\left(\frac{1}{2}+\frac{1}{2C_\epsilon}\right)\\
    = & \left(-(1-\alpha)\exp\left(-\alpha\frac{\lconst-\theta}{\sigma}\right)+1\right)\left(\frac{1}{2}-\frac{1}{2C_\epsilon}\right)\\
    & + \frac{\alpha(1-\alpha)}{\sigma}\frac{1}{C_\epsilon(\uconst-\lconst)}\left(\frac{\sigma}{\alpha}\lconst+\frac{\sigma^2}{\alpha^2}\right)\exp\left(-\frac{\alpha}{\sigma}(\lconst-\theta)\right)\\
    & - \frac{\alpha(1-\alpha)}{\sigma}\frac{1}{C_\epsilon(\uconst-\lconst)}\left(\frac{\sigma}{\alpha}\uconst+\frac{\sigma^2}{\alpha^2}\right)\exp\left(-\frac{\alpha}{\sigma}(\uconst-\theta)\right)\\
    & +\frac{\alpha(1-\alpha)}{\sigma}\left(\frac{1}{2}-\frac{\uconst+\lconst}{2C_\epsilon(\uconst-\lconst)}\right)\frac{\sigma}{\alpha}
    (\exp(-\frac{\alpha}{\sigma}(\lconst-\theta))-\exp(-\frac{\alpha}{\sigma}(\uconst-\theta)))\\
    & + (1-\alpha)\exp\left(-\frac{\alpha}{\sigma}(\uconst-\theta)\right)\left(\frac{1}{2}+\frac{1}{2C_\epsilon}\right)\\
    = &  \frac{1}{2}-\frac{1}{2C_\epsilon}
    + \frac{(1-\alpha)\sigma}{\alpha}\frac{1}{C_\epsilon(\uconst-\lconst)}\exp\left(-\frac{\alpha}{\sigma}(\lconst-\theta)\right)
    - \frac{(1-\alpha)\sigma}{\alpha}\frac{1}{C_\epsilon(\uconst-\lconst)}\exp\left(-\frac{\alpha}{\sigma}(\uconst-\theta)\right).
\end{align*}
Its first and second derivatives are 
\begin{align*}
    \Psi'_\epsilon(\theta) =  \frac{1-\alpha}{C_\epsilon(\uconst-\lconst)}\left(\exp\left(-\frac{\alpha}{\sigma}(\lconst-\theta)\right)-\exp\left(-\frac{\alpha}{\sigma}(\uconst-\theta)\right)\right), \\
    \Psi''_\epsilon(\theta) =  \frac{\alpha(1-\alpha)}{\sigma C_\epsilon(\uconst-\lconst)}\left(\exp\left(-\frac{\alpha}{\sigma}(\lconst-\theta)\right)-\exp\left(-\frac{\alpha}{\sigma}(\uconst-\theta)\right)\right)
\end{align*}
Since $\theta\leq\lconst$ and $\uconst>\lconst$, $\Psi'_\epsilon(\theta)$ is positive, and $\Psi''_\epsilon(\theta)$ is positive.
Moreover, we have
\begin{align*}
    |\Psi'_\epsilon(\theta)|<\frac{1-\alpha}{C_\epsilon(\uconst-\lconst)}
    \quad \text{and} \quad
    |\Psi''_\epsilon(\theta)| < \frac{\alpha(1-\alpha)}{\sigma C_\epsilon(\uconst-\lconst)}.
\end{align*}

The last case is the case where $\theta\geq\uconst$.
\begin{align*}
    \Psi_\epsilon(\theta) = & \alpha\exp\left(\frac{1-\alpha}{\sigma}(\lconst-\theta)\right)\left(\frac{1}{2}-\frac{1}{2C_\epsilon}\right)\\
    & - \frac{\alpha(1-\alpha)}{\sigma}\frac{1}{C_\epsilon(\uconst-\lconst)}\left(\frac{\sigma}{1-\alpha}(\lconst-\theta)-\frac{\sigma^2}{(1-\alpha)^2}\right)\exp\left(\frac{1-\alpha}{\sigma}(\lconst-\theta)\right)\\
    & +\frac{\alpha(1-\alpha)}{\sigma}\frac{1}{C_\epsilon(\uconst-\lconst)}\left(\frac{\sigma}{1-\alpha}(\uconst-\theta)-\frac{\sigma^2}{(1-\alpha)^2}\right)\exp\left(-\frac{1-\alpha}{\sigma}(\uconst-\theta)\right)\\
    & +\frac{\alpha(1-\alpha)}{\sigma}\left(\frac{1}{2}+\frac{\theta}{C_\epsilon(\uconst-\lconst)}-\frac{\uconst+\lconst}{2C_\epsilon(\uconst-\lconst)}\right)\frac{\sigma}{1-\alpha}(-\exp(\frac{1-\alpha}{\sigma}(\lconst-\theta))+\exp(\frac{1-\alpha}{\sigma}(\uconst-\theta)))\\
    & + \left(1-\alpha\exp\left(\frac{1-\alpha}{\sigma}(\uconst-\theta)\right)\right)\left(\frac{1}{2}+\frac{1}{2C_\epsilon}\right)\\
    = & \alpha\exp\left(\frac{1-\alpha}{\sigma}(\lconst-\theta)\right)\left(\frac{1}{2}-\frac{1}{2C_\epsilon}\right)\\
    & - \frac{\alpha(1-\alpha)}{\sigma}\frac{1}{C_\epsilon(\uconst-\lconst)}\left(\frac{\sigma}{1-\alpha}\lconst-\frac{\sigma^2}{(1-\alpha)^2}\right)\exp\left(\frac{1-\alpha}{\sigma}(\lconst-\theta)\right)\\
    & +\frac{\alpha(1-\alpha)}{\sigma}\frac{1}{C_\epsilon(\uconst-\lconst)}\left(\frac{\sigma}{1-\alpha}\uconst-\frac{\sigma^2}{(1-\alpha)^2}\right)\exp\left(-\frac{1-\alpha}{\sigma}(\uconst-\theta)\right)\\
    & +\frac{\alpha(1-\alpha)}{\sigma}\left(\frac{1}{2}-\frac{\uconst+\lconst}{2C_\epsilon(\uconst-\lconst)}\right)\frac{\sigma}{1-\alpha}(-\exp(\frac{1-\alpha}{\sigma}(\lconst-\theta))+\exp(\frac{1-\alpha}{\sigma}(\uconst-\theta)))\\
    & + \left(1-\alpha\exp\left(\frac{1-\alpha}{\sigma}(\uconst-\theta)\right)\right)\left(\frac{1}{2}+\frac{1}{2C_\epsilon}\right)\\
    = & \frac{1}{2} +\frac{1}{2C_\epsilon} + \frac{\alpha\sigma}{1-\alpha}\frac{1}{C_\epsilon(\uconst-\lconst)}\exp\left(\frac{1-\alpha}{\sigma}(\lconst-\theta)\right)
    -\frac{\alpha\sigma}{1-\alpha}\frac{1}{C_\epsilon(\uconst-\lconst)}\exp\left(-\frac{1-\alpha}{\sigma}(\uconst-\theta)\right).
\end{align*}
Its first and second derivatives are 
\begin{align*}
    \Psi_\epsilon'(\theta) = \frac{\alpha}{C_\epsilon(\uconst-\lconst)}\left(-\exp\left(\frac{1-\alpha}{\sigma}(\lconst-\theta)\right)+\exp\left(\frac{1-\alpha}{\sigma}(\uconst-\theta)\right)\right), \\
    \Psi_\epsilon''(\theta)
    = \frac{\alpha(1-\alpha)}{\sigma C_\epsilon(\uconst-\lconst)}
    \left(\exp\left(\frac{1-\alpha}{\sigma}(\lconst-\theta)\right)-\exp\left(\frac{1-\alpha}{\sigma}(\uconst-\theta)\right)\right).
\end{align*}
Since $\theta\geq\uconst$ and $\uconst>\lconst$, $\Psi'_\epsilon(\theta)$ is positive, and $\Psi_\epsilon''(\theta)$ is negative.
Moreover, since $(1-\alpha)(\lconst-\theta)/\sigma<(1-\alpha)(\uconst-\theta)/\sigma\leq 0$, we have
\begin{align*}
    |\Psi_\epsilon'(\theta)|< \frac{\alpha}{C_\epsilon(\uconst-\lconst)}
    \quad \text{and} \quad
    |\Psi_\epsilon''(\theta)|<\frac{\alpha(1-\alpha)}{\sigma C_\epsilon(\uconst-\lconst)}.
\end{align*}

We also analyze their behavior on the boundaries.
$\Psi_\epsilon(\theta)$ is continuous at $\theta=\lconst$ and $\uconst$ if and only if $\lim_{\theta\downarrow \uconst}\Psi_\epsilon(\theta) = \lim_{\theta\uparrow \uconst}\Psi_\epsilon(\theta)$ and $\lim_{\theta\downarrow \lconst}\Psi_\epsilon(\theta) = \lim_{\theta\uparrow \lconst}\Psi_\epsilon(\theta)$.
As we see below, these equations hold.
\begin{align*}
    \lim_{\theta\downarrow \uconst}\Psi_\epsilon(\theta) = \lim_{\theta\uparrow \uconst}\Psi_\epsilon(\theta) =  \frac{1}{2} +\frac{1}{2C_\epsilon} + \frac{\alpha\sigma}{1-\alpha}\frac{1}{C_\epsilon(\uconst-\lconst)}\exp\left(\frac{1-\alpha}{\sigma}(\lconst-\uconst)\right)
    -\frac{\alpha\sigma}{1-\alpha}\frac{1}{C_\epsilon(\uconst-\lconst)},\\
    \lim_{\theta\downarrow \lconst}\Psi_\epsilon(\theta) = \lim_{\theta\uparrow \lconst}\Psi_\epsilon(\theta) = 
    \frac{1}{2}-\frac{1}{2C_\epsilon}
    + \frac{(1-\alpha)\sigma}{\alpha}\frac{1}{C_\epsilon(\uconst-\lconst)}
    - \frac{(1-\alpha)\sigma}{\alpha}\frac{1}{C_\epsilon(\uconst-\lconst)}\exp\left(-\frac{\alpha}{\sigma}(\uconst-\lconst)\right).
\end{align*}
We next evaluate the existence of first and second derivatives at $\theta=\lconst$ and $\uconst$.
\begin{align*}
    \lim_{\theta\downarrow \uconst}\Psi'_\epsilon(\theta) = \lim_{\theta\uparrow \uconst}\Psi'_\epsilon(\theta) = 
    \frac{\alpha}{C_\epsilon(\uconst-\lconst)}\left(-\exp\left(\frac{1-\alpha}{\sigma}(\lconst-\uconst)\right)+1\right),\\
    \lim_{\theta\downarrow \lconst}\Psi'_\epsilon(\theta) = \lim_{\theta\uparrow \lconst}\Psi'_\epsilon(\theta) =
    \frac{1-\alpha}{C_\epsilon(\uconst-\lconst)}\left(1-\exp\left(-\frac{\alpha}{\sigma}(\uconst-\lconst)\right)\right).
\end{align*}
\begin{align*}
    \lim_{\theta\downarrow \uconst}\Psi''_\epsilon(\theta) = \lim_{\theta\uparrow \uconst}\Psi''_\epsilon(\theta) =
    \frac{\alpha(1-\alpha)}{C_\epsilon(\uconst-\lconst)}
    \left(\exp\left(\frac{1-\alpha}{\sigma}(\lconst-\uconst)\right)-1\right),\\
    \lim_{\theta\downarrow \lconst}\Psi''_\epsilon(\theta) = \lim_{\theta\uparrow \lconst}\Psi''_\epsilon(\theta) =
    \frac{\alpha(1-\alpha)}{\sigma C_\epsilon(\uconst-\lconst)}\left(1-\exp\left(-\frac{\alpha}{\sigma}(\uconst-\lconst)\right)\right).
\end{align*}

\section{COMPARISON WITH NON-PRIVATE ESTIMATOR}
\label{app:compare}

For comparison with existing work, we also consider the correct model case.
\begin{assumption}\label{asm:ald}
    Given $\mathbf{x}\in\Real^d$, $Y$ is a random variable sampled from the asymmetric Laplace distribution $f(\cdot;\alpha,\beta^{\top}\mathbf{x},\sigma)$, which is defined in (\ref{eq:asymmetric_laplace_distribution}).
    For each $i\in[n]$, $y_i$ is a realization of random variable $Y_i$ that is a copy of $Y$.
\end{assumption}
Under this condition, \cref{cor:quantile_normal_public} is more specified.
\begin{corollary}\label{cor:normal_quantile_public}
    Suppose \cref{asm:fyx_measurable,asm:FXlogfX_exists,asm:negative_definite,asm:ald} hold.
    The MLE $\hat{\beta}_n$ is  distributed asymptotically normally as $\sqrt{n}(\hat{\beta}_n-\beta^*) \to \mathcal{N}(0_d, I_{\beta^*}^{-1})$ where $I_{\beta^*}
    = F_X{\frac{\Psi_\epsilon'(\ip{\beta}{\mathbf{X}})^2}{\Psi_\epsilon(\ip{\beta}{\mathbf{X}})(1-\Psi_\epsilon(\ip{\beta}{\mathbf{X}}))}\mathbf{X}\mathbf{X}^\top}$. 
\end{corollary}

To obtain an intuitive understanding of the result, we roughly compare the Fisher information matrix derived in \Cref{cor:normal_quantile_public} and the non-private Fisher matrix (\ref{eq:Fisher_non_private}), and analyze some extreme cases. First, we consider the concentrated case in which the scale parameter $\sigma$ is extremely small.
For a $\sigma$ sufficiently small that $\sigma\ll|(1-\alpha)(\lconst-\ip{\beta^*}{\mathbf{x}})|$ and $\sigma\ll|\alpha(\uconst-\ip{\beta^*}{\mathbf{x}})|$ for most $\mathbf{x}$, 
\begin{align*}
    & \Psi(\ip{\beta^*}{\mathbf{x}}) \approx 
    \frac{1}{2} +
    \left(-\frac{\alpha}{1-\alpha}+\frac{1-\alpha}{\alpha}\right)
    \frac{\sigma}{2C_\epsilon}
    + \frac{\ip{\beta^*}{\mathbf{x}}}{2C_{\epsilon}}
    \quad\text{and}\quad
    \Psi'(\ip{\beta^*}{\mathbf{x}})
    \approx
    \frac{1}{2C_\epsilon}.
\end{align*}

Thus, 
\begin{align*}
    \frac{\Psi'(\ip{\beta^*}{\mathbf{x}})^2}{\Psi(\ip{\beta^*}{\mathbf{x}})(1-\Psi(\ip{\beta^*}{\mathbf{x}}))}
    \approx & 
    \frac{1}{C_\epsilon^2 - \left(\frac{1-2\alpha}{\alpha(1-\alpha)}\sigma+\ip{\beta^*}{\mathbf{x}}\right)^2}
    \geq \frac{1}{C_\epsilon^2-\left( \frac{1-2\alpha}{\alpha(1-\alpha)}\sigma+\lconst\right)^2}.
\end{align*}
In comparing this with (\ref{eq:non_private_normal}), we can see that the Fisher information matrix of our LDP estimator is $\Omega\left(\epsilon^2\frac{\sigma^2}{\alpha(1-\alpha)}\right)$ times smaller than that of the non-private estimator as $\epsilon\downarrow 0$.
This lower bound agrees with the complexity of $\epsilon$ but is $\sigma^2/\alpha(1-\alpha)$ times lower.
Since we assumed that $\sigma$ is small, this gap can be large.
Although our MLE tends to lose more information regarding the structure of $f_{\beta^*}$ than an optimal MLE, it experiences minimum information loss due to perturbation for privacy.

We omit the comparison of the MLE of the regression coefficient with the private $\mathbf{X}$.
The Fisher information matrix strongly depends on the structure of the distribution of $\mathbf{X}$.
We have no informative comparison in this case.

\section{ADDITIONAL NUMERICAL EVALUATION}
\label{app:numerical}

In this section, we perform some additional numerical evaluations with the real data, which is the same data used in \cref{sec:numerical}.

We implemented our simulation in Python.%, which is a popular language.
The necessary packages are written in requirements.txt.
The main part is written in experiment.py.
We made the Jupyter notebook files corresponding to each numerical evaluation.
Visualization of the results is also in the Jupyter notebook files.
You can open these files and run the simulations on your Jupyter notebook or Jupyter Lab.

Our supplemental material does not contain the real data used in the numerical evaluation.
Before running our program, please download the data from {\url{https://archive.ics.uci.edu/ml/datasets/Gas+Turbine+CO+and+NOx+Emission+Data+Set}}.
Then, put them into the folder "data/emission/".

\subsection{Evaluation of Private $\mathbf{X}$}

Here, we observe the behavior of our QMLE for the private $\mathbf{X}$ scenario, which is described in \cref{ssec:quantile_private}.
% We use the same data as \cref{sec:numerical} and observe the Frobenius norm of covariance matrices.

Due to implementation needs, we have made some modifications to the description in the main part.
First, we made some changes to $\Phi(\beta, \prix)$.
Theoretically, $\Phi$ and $1-\Phi$ never take negative values.
However, we found that the value of $\Phi$ can exceed $1$ by a small amount due to rounding error. %s caused by calculations with a finite number of digits.
Then, $1-\Phi$ is negative, and the computation corrupts since the log function is inputted a negative value.
To avoid this undesirable situation, we multiplied $\Phi$ by $e^{-0.000001}$.

Second, we changed the domain of $\hat{F}_\mathbf{X}$ because no element of each $\mathbf{x}_i$ is in the interval $[-1, 1]$.
In the simulation, each user truncates the components of $\mathbf{X}_i$ into the intervals
$[5, 10], [1000, 1030], [70, 100], [4, 6], [20, 30], [1000, 1100], [530, 570], [130, 170]$, and $[10, 15]$.
We recommend that the curators should set the intervals with the help of experts when they use our algorithm in reality.

% epsilons = np.array([10., 25, 50, 100])
%     num_trials = 1000
%     ns = np.array([100, 1000, 10000])

\begin{figure}[t]
    \centering
    \includegraphics[scale=0.5]{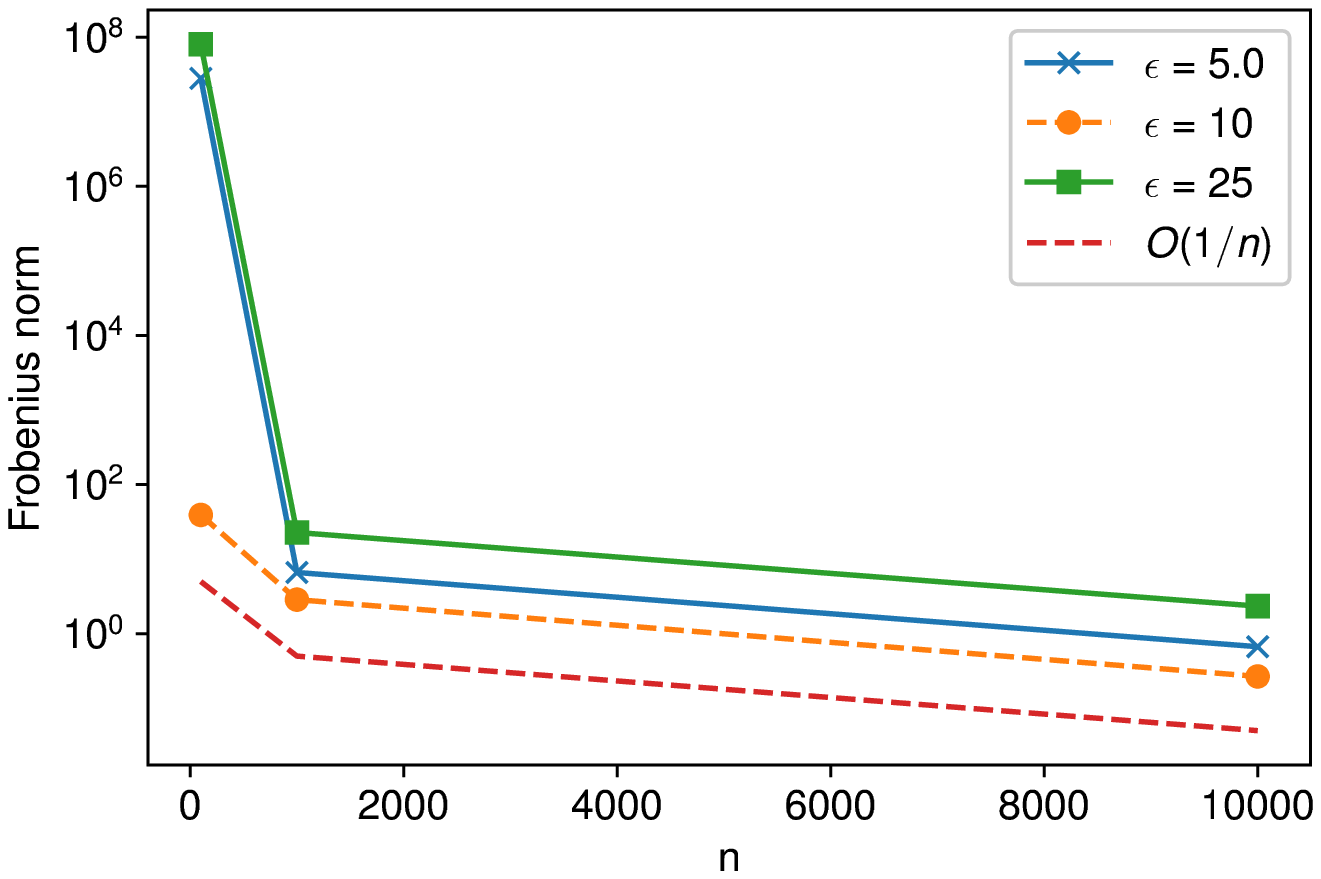}
    \caption{Frobenius norm of covariance matrices in private $\mathbf{X}$ scenario. The norm decrease in proportion to $1/n$ for each $\epsilon$.}
    \label{fig:appE1}
\end{figure}

We observe the covariance matrices for $n\in\{100, 1000, 10000\}$ and $\epsilon\in\{5.0, 10, 25\}$ with $\alpha=0.3$ and $\sigma=1.0$.
For each combination of $n$ and $\epsilon$, we sub-sample $n$ records $1,000$ times without replacement from the $36, 733$ records.
For each sub-data, we simulate the protocol described in \cref{ssec:quantile_private} and obtain a QMLE.
Then, we compute the Frobenius norm of covariance matrices of the $1,000$ QMLEs,

\cref{fig:appE1} shows the result.
The horizontal and vertical axes show $n$ and the value of each Frobenius norm in log-scale, respectively. 
For each $\epsilon$, with large $n$, the norm of the covariance matrix is smaller.
The decreasing speed is $O(1/n)$,
These properties are similar to those in the public $\mathbf{X}$ scenario, which is described in \cref{sec:numerical}.

\subsection{Evaluation of Effect of Truncation}

In this subsection, we evaluate the effect of the truncation in the public $\mathbf{X}$ scenario.

With $\epsilon = 2.5$ and $n = 10,000$, we try intervals $[50, 100], [40, 110], [30, 120]$ and $[20, 130]$ for the truncation.
The other setting is the same as \cref{sec:numerical}.

\begin{figure}[t]
    \centering
    \includegraphics[scale=0.5]{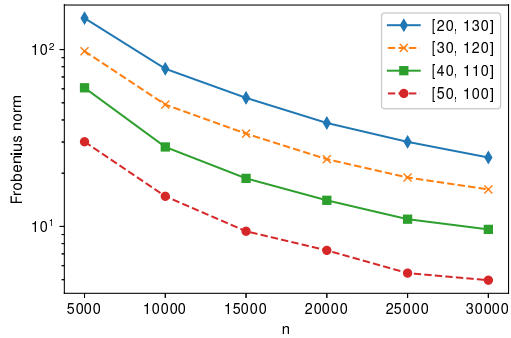}
    \caption{Frobenius norm of covariance matrices with various $[\lconst, \uconst]$s. A smaller interval makes the norm smaller.}
    \label{fig:appE2}
\end{figure}

\cref{fig:appE2} shows the result.
A shorter interval makes the estimators more concentrated.
We remark that the concentration does not necessarily imply a good approximation of the true distribution.
In general, there is a trade-off between bias and variance.

\subsection{Comparison with Non-private Estimator}

In this subsection, we evaluate the difference between the centers of the distributions of our QMLEs and the non-private QMLEs which is described in \cref{ssec:quantile}.
Our theoretical result does not say that those QMLEs converge to the same point.
Thus, we consider it with numerical simulations.

\begin{figure}[t]
    \centering
    \includegraphics[scale=0.45]{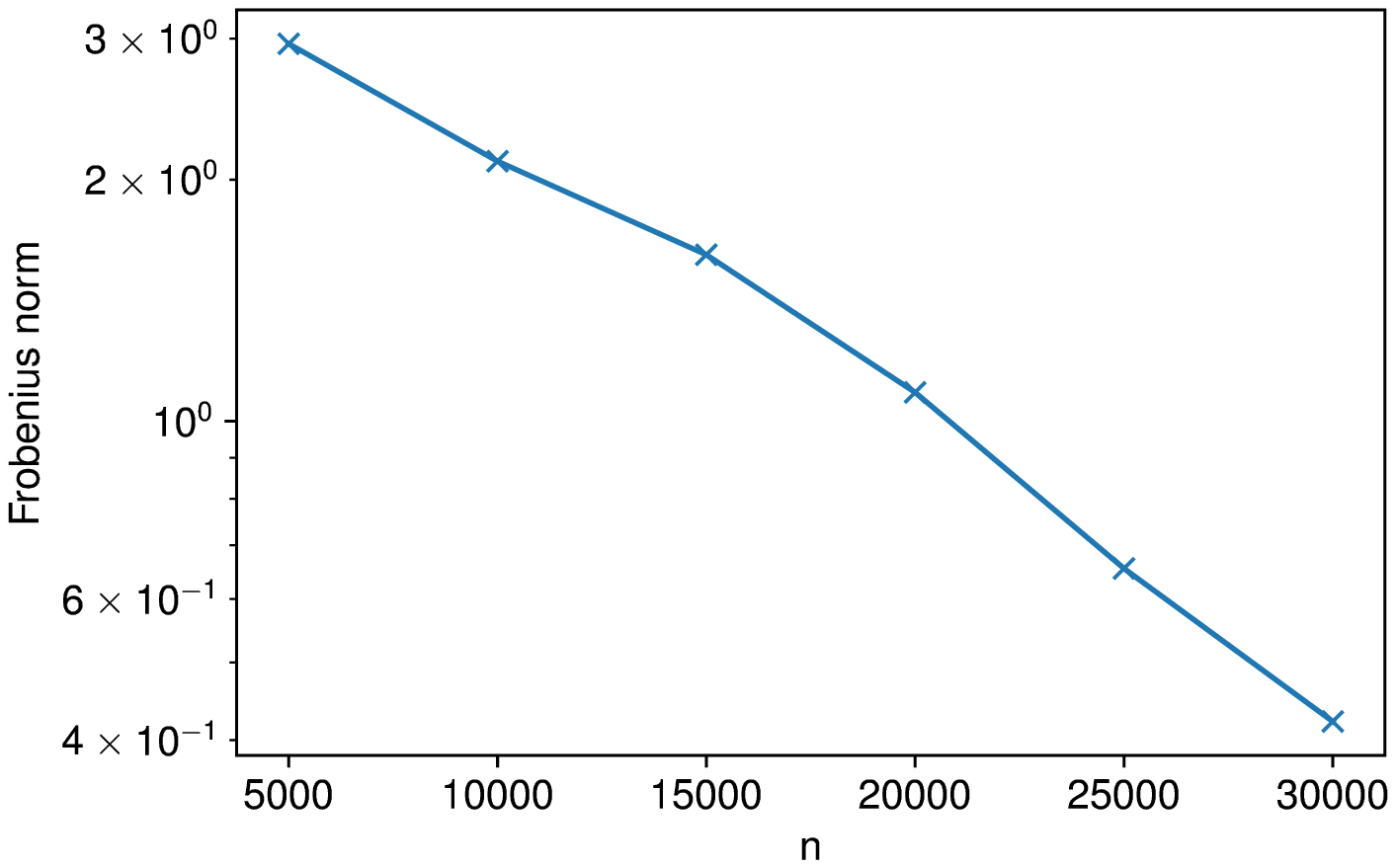}
    \caption{Frobenius norm of covariance matrices of non-private QMLE. It seems that the non-private QMLEs converge to one point.}
    \label{fig:appE3 non private}
\end{figure}

First, we observe the behavior of the non-private QMLE.
\cref{fig:appE3 non private} shows the Frobenius norm of covariance matrices.
It is seen that the non-private QMLEs converge to one point.
We treat the average vector of the non-private QMLEs with $n=30,000$ as the grand truth in the main observation as described below.
We remark that the "grand truth" can be biased.

We use the same simulation result used in \cref{sec:numerical}.
We compute the difference of the average vector of our QMLE and the grand truth and observe the norm for each $n$ and $\epsilon$.

\begin{figure}[t]
     \centering
     \includegraphics[scale=0.5]{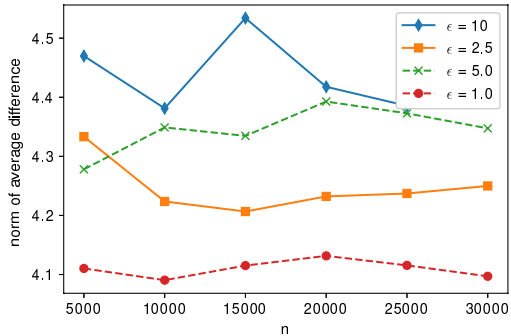}
     \caption{Norm of difference between the centers of non-private QMLEs and our QMLEs with various $n$ and $\epsilon$. The difference does not depend on $n$.}
     \label{fig:appE3 bias}
\end{figure}

\cref{fig:appE3 bias} shows the main result.
The horizontal and vertical axes show $n$ and the value of the norm of the covariance matrices, respectively.
The bias is not zero for all $\epsilon$.
Smaller $\epsilon$ tends to give smaller bias.
It is seen that $n$ does not affect the bias.
This result implies that the non-private QMLE and our QMLE can converge to different points.

\end{document}